\documentclass[runningheads]{llncs}
\usepackage{graphicx}

\usepackage{tikz}
\usepackage{comment}
\usepackage{amsmath,amssymb} 
\usepackage{color}
\usepackage{orcidlink}

\usepackage[accsupp]{axessibility}  

\usepackage{soul}
\usepackage{multirow}
\usepackage{booktabs}
\usepackage{makecell}
\usepackage{overpic}
\usepackage[export]{adjustbox}
\usepackage{subcaption}
\usepackage{algorithmic}
\usepackage{algorithm}

\def\ie{{i.e.}}

\def\etal{\emph{et al.}~}
\def\name{{MonoPix}~}
\def\namewo{{MonoPix}}

\newcommand{\mmc}[1]{\multicolumn{1}{l|}{#1}}

\newcommand{\mmcc}[1]{\multicolumn{1}{c|}{#1}}
\newcommand{\mmcn}[1]{\multicolumn{1}{l}{#1}}
\newcommand{\mmccn}[1]{\multicolumn{1}{c}{#1}}
\newcommand\tikzmark[2]{%
	\tikz[remember picture,baseline] \node[above, outer sep=0pt, inner sep=0pt] (#1){\phantom{#2}};%
}
\newcommand\link[2]{%
	\begin{tikzpicture}[remember picture, overlay, >=stealth, shift={(0,0)}, thick]
		\draw[very thick, ->] (#1) to (#2);
	\end{tikzpicture}%
}
\newcommand\linkdash[2]{%
	\begin{tikzpicture}[remember picture, overlay, >=stealth, shift={(0,0)}, thick]
		\draw[densely dashed, red, ultra thick] (#1) to (#2);
	\end{tikzpicture}%
}

\begin{document}
\pagestyle{headings}
\mainmatter
\def\ECCVSubNumber{2526}  

\title{Contrastive Monotonic Pixel-Level Modulation} 

\titlerunning{Contrastive Monotonic Pixel-Level Modulation} 
%
\author{Kun Lu\orcidlink{0000-0003-4698-9769} \and
	Rongpeng Li\orcidlink{0000-0003-4297-5060} \and
	Honggang Zhang\orcidlink{0000-0003-1492-1364}}
\authorrunning{K. Lu et al.}
%
\institute{Zhejiang University, Hangzhou, China \\
	\email{\{lukun199, lirongpeng, honggangzhang\}@zju.edu.cn}}


\maketitle

\begin{abstract}
	
	Continuous one-to-many mapping is a less investigated yet important task in both low-level visions and neural image translation. 
	In this paper, we present a new formulation called \namewo, an unsupervised and contrastive continuous modulation model, and take a step further to enable a pixel-level spatial control which is critical but can not be properly handled previously. The key feature of this work is to model the monotonicity between controlling signals and the domain discriminator with a novel contrastive modulation framework and corresponding monotonicity constraints. 
	We have also introduced a selective inference strategy with logarithmic approximation complexity and support fast domain adaptations. 
	The state-of-the-art performance is validated on a variety of continuous mapping tasks, including AFHQ cat-dog and Yosemite summer-winter translation. The introduced approach also helps to provide a new solution for many low-level tasks like low-light enhancement and natural noise generation, which is beyond the long-established practice of one-to-one training and inference. 
	Code is available at \url{https://github.com/lukun199/MonoPix}.

\end{abstract}

\section{Introduction}

Deep learning has significantly revolutionized the way we process and represent images. 
Though in most high-level tasks, we may safely assume that the input-ground truth pairs are deterministic and perfectly aligned \cite{He_2016_CVPR,Girshick_2015_ICCV,long2015fully} (e.g., class labels, bounding boxes, and key points), it does not hold true in many low-level scenarios. In low-light image enhancement \cite{lore2017llnet,chen2018learning} and natural noise generation \cite{chen2018image,kim2019grdn} for instance, the ever-changing environment makes it even impossible to provide firm ground truth pairs.
Directly adopting a one-to-one (O2O) mapping on these pseudo labels can result in several problems: in the training stage, it ignores the one-to-many (O2M) mapping property, and inherently invites certain divergence and instability; for evaluation and deployment purposes, it incurs large variance in different environments \cite{li2021low} and lacks certain scalability.

\begin{figure*}[b!]
	\centering
	\includegraphics[width=0.85\textwidth, valign=m]{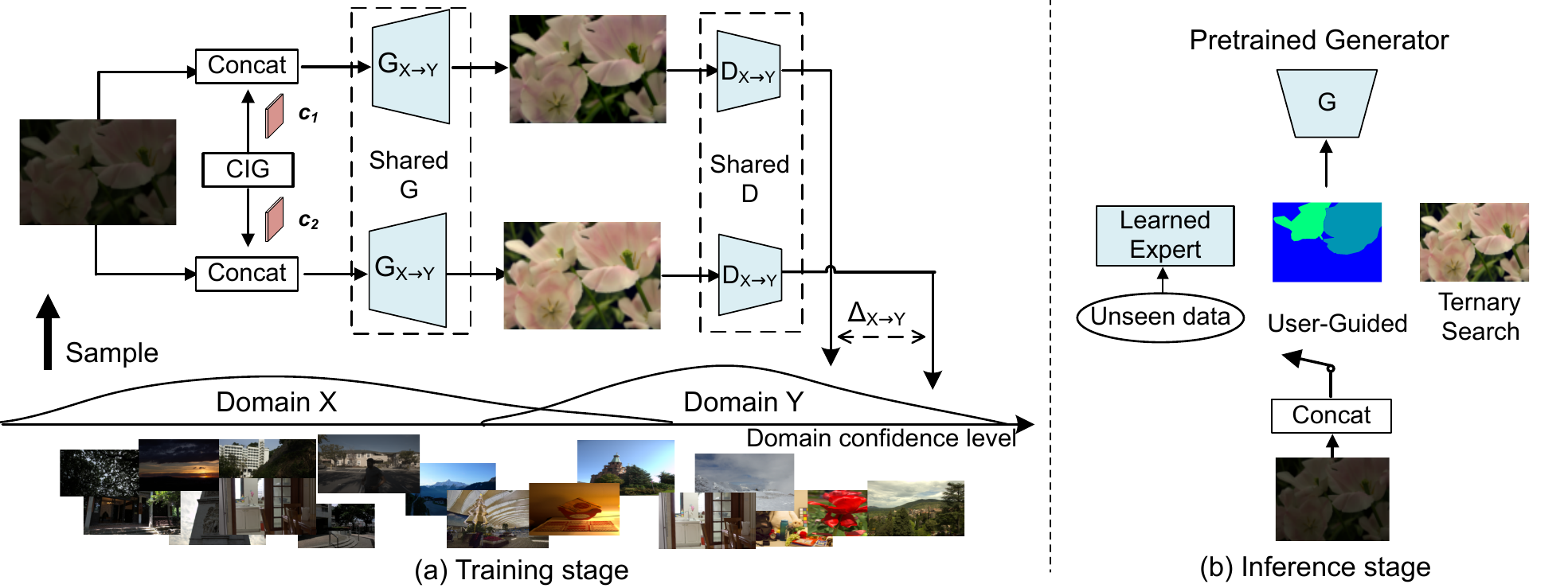}
	\caption{Overview of \name framework. Blue blocks are learnable parts. CIG denotes contrastive intensity generator (a procedure) that randomly produces two spatial control signals $\boldsymbol{c_1}$ and $\boldsymbol{c_2}$ for contrastive training. $\Delta_{X\mapsto Y}$ is the confidence difference. (a) During training, we model the monotonicity between control signal and domain confidence. (b) We enable multiple inference strategies or perform fast adaptation through another expert network}
	\label{fig:main}
\end{figure*}
To enable O2M mappings, a direct solution is to introduce conditional generative models \cite{jo2021deep,chen2020controllable}, or carry out a multi-branch ensemble on several sub-modules \cite{lu2020tbefn,ni2021controlling}. The major obstacle comes from the lack of dense and varied discrete image pairs to support the subsequent training. Some other methods approach continuous generation by interpolating between two models, or tuning an extra adaptation module \cite{he2019modulating,shoshan2019dynamic,wang2019deep,he2020interactive,lee2020smoother}. Nevertheless, methods in this category involve two correlated networks which are usually obtained from a fine-tuning or re-training. As a result, they not only prevent an end-to-end training, but also increase the storage burden and inference complexity. Moreover, all these methods can suffer from a training-inference gap that, a discrete training is not sufficient to provide a continuous modulation at inference stage. 

On the other hand, 
through the content and style disentanglement \cite{lee2018diverse,karras2019style,choi2018stargan}, it is shown that a single generative network is capable of handling varied styles and attributes in an end-to-end manner, and provides a smooth generation by style interpolation \cite{karras2019style,choi2020stargan}. 
To better secure the intermediate generation quality, works on continuous cross-domain translation further propose to constrain the in-between samples by coordinating two domain discriminators \cite{gong2019dlow,mao2022continuous}, or introducing some extra constraints \cite{wu2019relgan}. However, to obtain a desired output, 
a user either has to change the random style noise at an unpredictable time complexity, or is required to provide a proper reference style which may not always be plausible. 
In addition, these methods are primarily focused on an image-level holistic style translation, but not typically designed for many tasks where a pixel-level modulation is preferred.

These limitations pique our interest; and we therefore devote to providing a new solution that simultaneously enables an unsupervised, end-to-end, continuous, and monotonic pixel-level image control. 
By injecting a spatial controlling signal into the feature space, and benefiting from the inductive bias of convolutions, we render it possible to directly manipulate from a fine-grained image-like interface. The monotonicity control is inspired by the recent progress on contrastive learning \cite{ye2019unsupervised,chen2020simple} and image-to-image translation \cite{gong2019dlow,mao2022continuous}, where we further devise a contrastive modulation scheme that connects the control monotonicity with the confidence level of a domain discriminator. In this way, one image more ``resembles'' the target domain will get rewarded with a higher discriminator score; and the training could proceed without any other style level supervisions. A similar previous work CUT \cite{park2020contrastive} introduces patchwise contrastive loss (on different patches) for image translation, while it does not support image editing and O2M mapping. A schematic illustration of our approach can be found in Figure \ref{fig:main}. 
Simple though these ideas are, they construct a competitive new solution for continuous domain translation, typically those in conjunction with conventional low-level tasks.

We summarize the contributions of this work:
\begin{itemize}
	\item We investigate a new research area of continuous pixel-level image modulation which has not been particularly investigated before, and provide the corresponding solution. The surveyed topic enables an interesting unsupervised continuous translation and provides a fine-grained image editing tool.
	\item We propose a new contrastive learning-based solution for continuous image translation, which can be seamlessly integrated into existing GAN-based one-to-one mapping schemes, and is both training and inference-efficient. The proposed approach does not need any domain-specific labels and has no training and inference gap.
	\item We provide several basic evaluation metrics for continuous generation, and validate the state-of-the-art performance of the proposed model on both the translation continuity and quality. Extensive results on low-level tasks also demonstrate the effectiveness of our proposed approach.
	\item We are one of the first that revisit the long-established O2O mapping protocol in many low-level tasks. Our work suggests it is plausible and probably more flexible to formulate them in an O2M or M2M manner instead.
\end{itemize}

\section{Related Work}

\noindent\textbf{One-to-Many Mappings in Low-Level Task.} 
O2M mapping commonly occurs in low-level image processing. It is evidenced in recent literature that many tasks, such as deraining \cite{ni2021controlling}, noise modeling \cite{chen2020controllable,he2019modulating,he2020interactive}, and low-light image enhancement \cite{klyuchka2020cel,sun2021enhance,jo2021deep} can be categorized into this scope. 
Data-guided approaches model the diversity with synthesized one-to-many pairs (usually with physical models) and generate them with the corresponding control signals \cite{jo2021deep,chen2020controllable,he2020interactive,cai2021toward,pumarola2018ganimation}.
As an alternative to the dense training pairs, image modulation methods address this task in a two-stage training manner. 
Upchurch \etal \cite{upchurch2017deep} propose to train two networks and implement feature interpolations. In CFSNet \cite{wang2019cfsnet} the extra network is further simplified into only a tuning branch. 
Inspired by the normalization operations, He \etal \cite{he2019modulating} propose to insert a per-layer modulation module and adjust them when trained on another enhancing level. Similarly, DynamicNet \cite{shoshan2019dynamic} performs adaptation with several extra tuning blocks. 
Besides feature interpolation, it is also shown that modulations can take place via interpolation on the network parameters \cite{wang2019deep} or image residuals \cite{he2020interactive}. 
There are also some efforts that attempt to perform consecutive enhancement \cite{zamir2021multi,ren2019progressive}, but the increased complexity and lack of control range also limit their applications.

In contrast, we present a continuous modulation in a totally unsupervised\footnote{In this paper, we typically refer to methods that perform domain translation without the need for paired images and intermediate domain intensity labels.}, end-to-end, and inference-efficient manner by training a single network that executes all these tasks. There is no training and inference gap in the proposed approach, which by design allows a continuous and realistic generation. 

\noindent\textbf{Image-to-Image Translation.} 
CNN-based image translation mainly takes the style representation as an auxiliary loss \cite{gatys2015texture,li2016combining,johnson2016perceptual,simonyan2014very}. 
Different from designing explicit style descriptors, it is observed by Dumoulin \etal \cite{dumoulin2016learned} that incorporating an affine transformation after the instance normalization (IN) \cite{ulyanov2016instance} layer can lead to a conditional style generation, which further inspires data-driven style manipulation methods like AdaIN \cite{huang2017arbitrary} and WCT \cite{li2017universal}. StyleBank \cite{chen2017stylebank}, on the other hand models the style translation with a trainable bank that disentangles the style representation. It is also shown in \cite{zhang2018multi} and \cite{li2017diversified} that by injecting the style information as part of the input, the model can generate varied samples with the same set of parameters. 

As another line of work, GAN \cite{goodfellow2014generative} provides an important tool for image translation. Pioneering works like Pix2Pix \cite{isola2017image} and CycleGAN \cite{zhu2017unpaired} introduce adversarial and unsupervised conditional generation, but are O2O mapping models. To enable a diverse translation, non-one-to-one mappings are thereafter investigated \cite{choi2018stargan,lu2018attribute,huang2018multimodal,almahairi2018augmented}. As one of the most popular and successful schemes, disentangling the content and style representation exhibits a great success on a wealth of attribute and style translation tasks \cite{lee2018diverse,huang2018multimodal,choi2020stargan,yu2019multi,karras2019style}, and enables a continuous translation by interpolating between two latent vectors \cite{karras2019style,choi2020stargan}. 
Recently, works on continuous cross-domain translation further refine the quality of intermediate images by introducing an interpolation discriminator \cite{wu2019relgan,lira2020ganhopper}, constraining the intermediate results with discriminators from both sides \cite{gong2019dlow,mao2022continuous}, or by exploiting the path of interpolation and translation manifold \cite{chen2019homomorphic,pizzati2021comogan,liu2021smoothing}. 

Different from these existing works, the proposed approach eschews the need for either any extra discriminator or a weighted bi-lateral supervision that can hardly scale to model many-to-one mappings. Instead, we teach one discriminator by contrastively feeding two samples and modeling the consistency between control signals and translation monotonicity. 

\noindent\textbf{Image Editing.}
Our work is also related to image editing. This task can be achieved generally by learning a conditional image-level mapping \cite{isola2017image,zhu2017unpaired,lee2020maskgan,lv2021learning} or by exploring the latent semantic trajectories \cite{Jahanian2020On,zhu2020domain,collins2020editing,Shen_2021_CVPR,zhuang2021enjoy}. Modern image editing methods prove to be effective in modeling multiple image-level features such as semantic contents, geometric shapes, and styles \cite{shi2021learning,ling2021editgan}; but there are a few that provide a spatial control \cite{kim2021exploiting,zhu2020sean,park2019semantic}, and typically in an unsupervised manner. The proposed method, further, provides a fine-grained, unsupervised, and spatial-level editing, whilst exhibiting both fidelity to inputs and a strong scalability for style and content manipulation.

\section{Continuous Cross-Domain Pixel-Level Control}
First we formalize a brief description of the newly examined task. Consider two domains $X$ and $Y$ each with corresponding image samples i.e. ${\{\boldsymbol{x_i}\}}^M_{i=1}$ and ${\{\boldsymbol{y_i}\}}^N_{i=1}$, together with a control signal $\boldsymbol{c}$ of the shape $W \times H$ representing the translation intensity of each pixel. We learn a mapping function $G_{X\mapsto Y}: X \times C \mapsto Y$ that translate from $X$ to $Y$, and from $Y$ to $X$ if necessary. Moreover, the translation is required to be monotonic. Thus, for each image or patch $\boldsymbol{x_i}$ and two control signals $\boldsymbol{c_1}$ and $\boldsymbol{c_2}$, we require the translated result $G_{X\mapsto Y}(\boldsymbol{x_i}, \boldsymbol{c_2})$ better belongs to the target domain $Y$ than its counterpart if $c_2^{i,j} > c_1^{i,j}$ for each pixel position. 
In the following, we will elaborate on the detailed framework and corresponding loss functions that lead to accomplishing this goal. 

\subsection{Pixel-Level Contrastive Generation}
In existing works on unsupervised domain translation, the conditional style vector 
is usually constructed as a flat one \cite{karras2019style,choi2018stargan} or with coarse spatial shapes \cite{kim2021exploiting}. 
These approaches are moderately satisfactory when controlling the local attributes, but can not enable a fine-grained control. 

To tackle this problem, we first visit the pixel-level control signal. This idea is plain and easy to implement, where we can directly 
formulate the control signal as an extra input channel. 
Without loss of generality, we always assume that the pixel-level control signal is bounded in $[0, 1]$. Then, the translated result $\hat{\boldsymbol{y}}$ of an input $\boldsymbol{x}$ could be represented as (and vice versa from $Y$ to $X$)

\begin{equation}
	\hat{\boldsymbol{y}} = G_{X\mapsto Y}(\texttt{concat}(\boldsymbol{x}, \boldsymbol{c})) \quad  \text{for each} \quad c^{i,j} \in [0, 1]
	\label{equ:concat}
\end{equation}

However, to enable such a pixel-level training is non-trivial. First, modern training schemes and datasets are only focused on holistic attributes, but there are no pixel-level labels. Second, we do not have a gallery of domain-specific intermediate labels that guide the modulation process, except for the coarse binary tags - 0 or 1, which can not model the domain distribution and precludes conventional absolute intensity-based training. 
To overcome these problems, we propose to formulate a contrastive learning-based approach instead, that trains the domain generator $G$ by relative intensity. In the training stage, we first randomly generate two pixel-level control signals $\boldsymbol{c_1}$ and $\boldsymbol{c_2}$ for each input patch, then we encourage the discriminator to yield a higher domain-specific confidence level for that with a higher translation strength. Since the discriminator $D$ is trained to examine both the quality and domain-specific features of generated images, a higher confidence level for $D$ means a better likelihood that the output belongs to the target domain. This is different from DLOW \cite{gong2019dlow}, who needs to train two discriminators and overlooks the underlying data distributions in the source domain. Note that in the training stage, we empirically set a control signal $\boldsymbol{c}$ as one filled with the same value (i.e., $c^{1,1}=c^{1,2}=...=c^{W,H}$), considering the randomness of control signal and the inductive bias of convolutional operation. On the contrary, assigning each $c^{i,j}$ with randomly generated numbers during training is tedious, and also violates the smoothness of neighboring pixels.

\subsection{Normalization-Compatible Translator} 
\label{sec:InsNorm}
Although concatenating the pixel-level control signal $\boldsymbol{c}$ can be direct and simple, it is incompatible with commonly used instance normalization modules \cite{ulyanov2016instance}. Denoted by $\{\boldsymbol{\mathcal{K}}^{i}\}_{i=1}^C$ a convolution kernel where $i$ represents the \emph{i-th} channel, $\mathtt{Idx}$ the channel index function and $\otimes$ the convolution operation, the convoluted result for $\texttt{concat}(\boldsymbol{x}, \boldsymbol{c})$ can be calculated as $\boldsymbol{x} \otimes \boldsymbol{\mathcal{K}}^{i \neq \mathtt{Idx}(\boldsymbol{c})} + \boldsymbol{c} \otimes \boldsymbol{\mathcal{K}}^{i=\mathtt{Idx}(\boldsymbol{c})}$. As a result, the numerical difference between two contrastive samples is expected to be $(\boldsymbol{c_2}-\boldsymbol{c_1}) \otimes \boldsymbol{\mathcal{K}}^{i=\text{Idt}(\boldsymbol{c})}$, which becomes a constant when $\boldsymbol{c}$ is element-wise the same (in convolutions, a kernel is shared across the spatial dimension). 
It could also be concluded that any linear operation before instance normalization would cause the same problem (for example adding the broadcast control signal to the input) as the constant difference on two channels would be eliminated by IN. To prevent this, we introduce a simple non-linear transformation like leaky ReLU \cite{maas2013rectifier} when adopting instance normalization. Visualizations on the activated 
domain-specific features (by identifying the positions where a non-linear transformation takes place) can be found in the supplement. Batch normalization, on the other hand, introduces certain contrastivity inside the training samples but does not impede the overall monotonicity.

\subsection{Objective Functions}
Based on the proposed contrastive pixel-level modulation framework, we provide the corresponding constraints steering towards a monotonic and cross-domain control. Since the translation can be bi-directional, we will always take that from $X$ to $Y$ as an example. 

\noindent\textbf{Monotonicity Loss.}
We employ the output value of target discriminator to model the likelihood of belonging to this domain. Given an input image $\boldsymbol{x}$ and two pixel-level control signal $\boldsymbol{c_1}$ and $\boldsymbol{c_2}$ where we assume $c_2^{i,j} > c_1^{i,j}$, the confidence difference $\boldsymbol{\Delta^{tar}_{X\mapsto Y}}$ between translated results can be calculated as:

\begin{equation}
	\boldsymbol{\Delta^{tar}_{X\mapsto Y}} = D_Y(G_{X\mapsto Y}(\boldsymbol{x},\boldsymbol{c_2}))
	- D_Y(G_{X\mapsto Y}(\boldsymbol{x},\boldsymbol{c_1}))
	\label{equ:Delta}
\end{equation}

\begin{figure*}[b]
	\centering
	\begin{minipage}[b]{\textwidth}
		\begin{minipage}[b]{0.45\textwidth}
			\centering
			\includegraphics[width=10em]{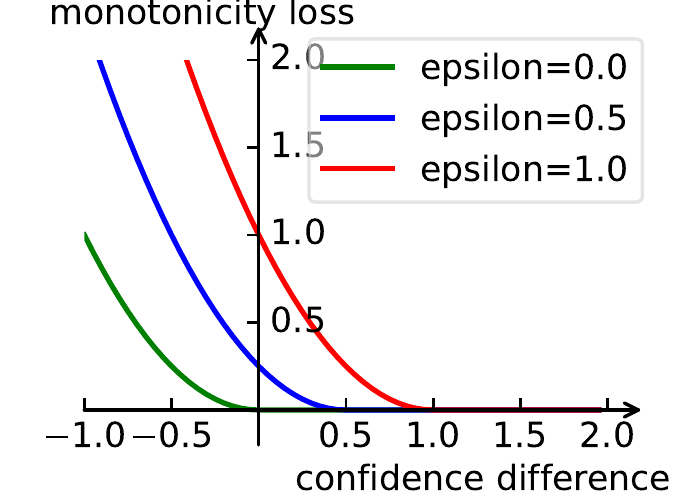}
			\caption{Example curves of the monotonicity loss. A positive $\epsilon$ encourages a margin on confidence difference}
			\label{fig:loss_mono}
		\end{minipage}
		\hfil
		\begin{minipage}[b]{0.45\textwidth}
			\centering
			\includegraphics[width=10em]{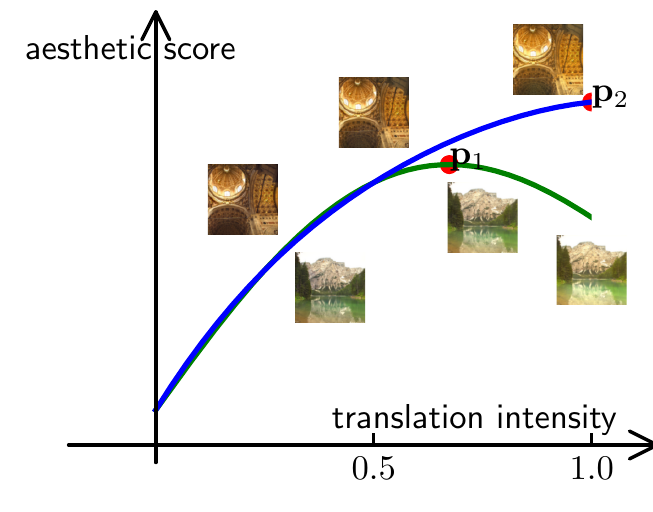} 
			\caption{Illustration of the aesthetic gap. $\boldsymbol{p}$ denotes the best balance point 
			}
			\label{fig:ts_demo}
		\end{minipage}
	\end{minipage}
\end{figure*}

Under the monotonicity constraint $c_2^{i,j} > c_1^{i,j}$, we encourage the confidence difference $\boldsymbol{\Delta^{tar}_{X\mapsto Y}}$ to be positive for any $\boldsymbol{c_1}$ and $\boldsymbol{c_2}$:

\begin{equation}
	\mathcal{L}_{mono-X2Y} = \mathbb{E}_{\boldsymbol{x}\sim X, \boldsymbol{c_1}, \boldsymbol{c_2} \sim C} \Vert \max \left( \epsilon - \boldsymbol{\Delta^{tar}_{X\mapsto Y}}, 0 \right) \Vert_2^2
	\label{equ:mono}
\end{equation}
where $\epsilon > 0$ is a hyperparameter controlling the strength of monotonicity, which we call a \emph{margin} here. As illustrated in Figure \ref{fig:loss_mono}, the above loss exerts a quadratic punishment when $\boldsymbol{\Delta^{tar}_{X\mapsto Y}} < \epsilon$, and thus introduces not only a positive $\boldsymbol{\Delta^{tar}_{X\mapsto Y}}$, but also a positive margin. Omitting $\epsilon$ can cause a trivial solution where the confidence difference is only a little greater than zero and finally a weak control.

\label{sec:domainfidelity}
\noindent\textbf{Domain Fidelity Loss.}
Although monotonicity loss enables a continuous cross-domain control, there exists an asymmetry issue. As we know that a well-trained discriminator can only produce a high confidence level for a \emph{real} image, constraining the confidence level to be low, as in the case of $\boldsymbol{c_1}$, may not be promising. The discrepancy can lead to a biased control that lacks in the diversity and control quality. 
Though this can be mitigated by applying a weak paired loss (e.g., \emph{L1} or \emph{MSE}) between the input image and the translated one, a rigid identity mapping also introduces certain instability for $\boldsymbol{c_2}$ and involves much parameter tuning. As a surrogate, we not only require a higher confidence level for $\boldsymbol{c_2}$ from the target domain discriminator, but also a higher confidence level for $\boldsymbol{c_1}$ from the source discriminator. We formulate the domain fidelity loss as:

\begin{equation}
	\mathcal{L}_{df-X2Y} = \mathbb{E}_{\boldsymbol{x}\sim X, \boldsymbol{c_1}, \boldsymbol{c_2} \sim C} \Vert \max \left(\epsilon - \boldsymbol{\Delta^{src}_{X\mapsto Y}}, 0 \right) \Vert_2^2
	\label{equ:df}
\end{equation}
where $\boldsymbol{\Delta^{src}_{X\mapsto Y}}$ is defined as:

\begin{equation}
	\boldsymbol{\Delta^{src}_{X\mapsto Y}} = D_X(G_{X\mapsto Y}(\boldsymbol{x},\boldsymbol{c_1}))
	- D_X(G_{X\mapsto Y}(\boldsymbol{x},\boldsymbol{c_2}))
	\label{equ:DeltaPrime}
\end{equation}

\noindent\textbf{GAN Loss.}
\name also utilizes the adversarial loss \cite{goodfellow2014generative} and cycle consistency loss \cite{zhu2017unpaired} to provide a proper gradient when training the bi-directional translation network. GAN loss is specialized in constraining the overall quality:

\begin{equation}
	\min \max \ \mathcal{L}_{adv-X2Y} = \mathbb{E}_{\boldsymbol{y}\sim Y} [D_Y(\boldsymbol y)] + 
	\mathbb{E}_{\boldsymbol{x}\sim X, \boldsymbol{c}\sim C}[1 - D_Y(\hat{\boldsymbol{y}})]
	\label{equ:adv}
\end{equation}

We adopt cycle consistency loss since in many low-level tasks we are more concentrated on the intensity and stability of enhancement instead of the diversity of appearance. To this end, we expect that the input image could be recovered from a reverse translation with the same intensity $\boldsymbol{c}$:

\begin{equation}
	\mathcal{L}_{cyc-X2Y} = \mathbb{E}_{\boldsymbol{x}\sim X} \left[ \Vert G_{Y\mapsto X}(G_{X\mapsto Y}(\boldsymbol x, \boldsymbol c), \boldsymbol c) - \boldsymbol x \Vert_1 \right]
	\label{equ:cyc}
\end{equation}

\noindent\textbf{Overall Objective.}
To keep the presentation succinct, we take the translation from $X$ to $Y$ as an example when constructing these loss functions. Similarly losses in the reverse direction could also be formulated. 
For the generator $G=\{G_{X\mapsto Y}, G_{Y\mapsto X}\}$ and discriminator $D=\{D_X, D_Y\}$, the overall training objective is expressed as:

\begin{equation}
	\begin{aligned}
		&\min G \quad \mathcal{L}_{G} = \mathcal{L}_{adv} + \lambda_{cyc} \mathcal{L}_{cyc} + 
		\lambda_{mn} \mathcal{L}_{mono} + 
		\lambda_{df} \mathcal{L}_{df} \\ 						  
		&\min D \quad \mathcal{L}_{D} = \mathcal{L}_{adv} + 
		\lambda_{mn} \mathcal{L}_{mono} + 
		\lambda_{df} \mathcal{L}_{df}
		\label{equ:lossALL}
	\end{aligned}
\end{equation}
where $\lambda_{cyc}$, $\lambda_{mn}$ and $\lambda_{df}$ are hyperparameters balancing different losses. 

\subsection{Selective Inference}
\name naturally generates a gallery of candidate results for a given input by constraining the monotonicity. However, a high translation intensity, and hence a high domain discriminator's confidence level does not necessarily yield a satisfactory result. As shown in Figure \ref{fig:ts_demo}, an image can much ``resemble'' the target domain but is inconsistent with human perception (aesthetic score). 

We assume there lies a perfect balance point $\boldsymbol{p}$ that yields the best aesthetic score, where this score can be any reasonable metric, such as human perception, and quantitative metrics like PSNR and SSIM \cite{wang2004image} when a reference image is available. 
To approximate $\boldsymbol{p}$, \name enables a variety of inference strategies with different execution complexity:

\noindent\textbf{Exhaustive Inference.}
Benefiting from the monotonicity, we can always sample $N$ control signals and choose the best result. The complexity is $O(N)$.

\label{sec:ternarysearch}
\noindent\textbf{Ternary Search.}
Since the translation intensity - aesthetic score curve can be a monotonically increasing one or consists of two continuous monotonic parts, it is \emph{unimodal-like}. Finding the local maxima can thus be efficiently solved by using \emph{ternary search}. The details of this method are described in the supplementary material. It has the $O(\log N)$ complexity.

\noindent\textbf{Guidance from Learned Expert.}
We can further decrease the complexity to $O(1)$ by introducing an expert network $P$ that mimics target distribution. It can be trained in a few-shot manner and even on unseen datasets for fast adaptation.

\section{Experiments and Results}
We carry out experiments on both the commonly used domain translation datasets AFHQ cat-dog \cite{choi2020stargan} and Yosemite summer-winter \cite{zhu2017unpaired}, and low-level tasks including prevalent LOL low-light enhancement \cite{wei2018deep} and SIDD noise generation \cite{abdelhamed2018high}. For a fair comparison, we will always report the results when $\boldsymbol{c}$ is set to be element-wise the same when not specified.

\subsection{Metrics}
For continuous cross-domain image generation, we evaluate both the dynamic translation process and the quality of translated images:

\noindent\textbf{AL/Rg:} 
As we expect that an image should change monotonically from the input one, the foremost concern comes from the absolute linearity (AL) and control range (Rg). For an input image, we first modulate it by varying the intensity level $\boldsymbol{c}$ from 0 to 1 with an interval of 0.1. 
Then, AL score can be evaluated by the \emph{Pearson correlation coefficient} between these intensity levels and the LPIPS distances \cite{zhang2018unreasonable} against the input image. Rg score is calculated by the $range$ of these values, which reflects both the fidelity and difference. 

\noindent\textbf{RL/Sm:} 
Besides absolute linearity, we follow \cite{wu2019relgan} to measure the smoothness of adjacent intermediate pairs. To both overcome the drawback of \emph{standard deviation} as used in \cite{wu2019relgan} (which is sensitive to the absolute input value), and keep in comparison with AL, we calculate relative linearity (RL) from the linearity of cumulative adjacent LPIPS distances, where a high RL score indicates the translation is even between adjacent samples. The smoothness (Sm) can be evaluated by measuring the maximum value of these adjacent LPIPS distances. 

\noindent\textbf{ACC/FID:} 
Following \cite{mao2022continuous,wu2019relgan}, we adopt a pre-trained ResNet \cite{He_2016_CVPR} binary classifier to measure the success of translation. For each image, we report the highest confidence level of the classifier among the consecutive modulation process. The quality of translation can be evaluated by Fréchet inception distance (FID) \cite{heusel2017gans} between the whole trajectory (with interval of 0.1) and real images, as in \cite{liu2021smoothing}.

We use the corresponding paired evaluation metrics in low-level tasks. For low-light image enhancement, we adopt PSNR, SSIM, and LPIPS. For blind noise generation, we leverage average KL divergence (AKLD) \cite{yue2020dual}. 

\begin{figure*}[b!]
	\centering
	\resizebox{\linewidth}{!}{
		\setlength{\tabcolsep}{0.003\linewidth}
		\begin{tabular}{c c c | c c c c c c c c c c c c c}
			\multicolumn{15}{c}{Summer $\mapsto$ Winter}\\
			\multirow{1}{*}[0.7cm]{\rotatebox{90}{\tiny{CycleGAN-DNI}}}
			& \includegraphics[width=4em, valign=m]{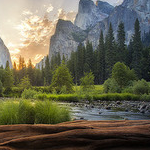}
			& ~
			& ~
			& \includegraphics[width=4em, valign=m]{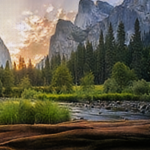}
			& \includegraphics[width=4em, valign=m]{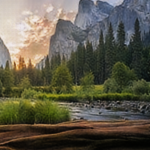}
			& \includegraphics[width=4em, valign=m]{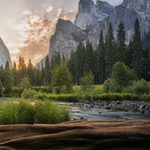}
			& \includegraphics[width=4em, valign=m]{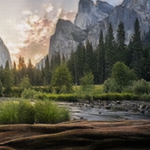}
			& \includegraphics[width=4em, valign=m]{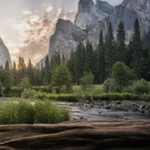}
			& \includegraphics[width=4em, valign=m]{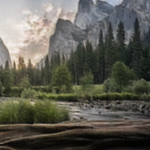}
			& \includegraphics[width=4em, valign=m]{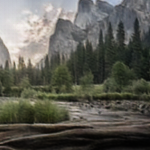}
			& \includegraphics[width=4em, valign=m]{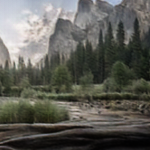}
			& \includegraphics[width=4em, valign=m]{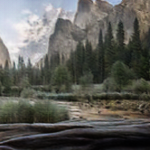}
			& \includegraphics[width=4em, valign=m]{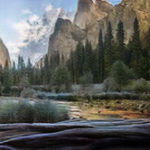}
			& \includegraphics[width=4em, valign=m]{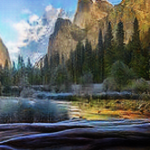}
			\\
			\multirow{1}{*}[0.35cm]{\rotatebox{90}{\tiny{StarGANv2}}}
			& \includegraphics[width=4em, valign=m]{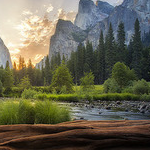}
			& ~ 
			& ~	
			& \includegraphics[width=4em, valign=m]{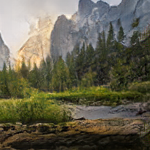}
			& \includegraphics[width=4em, valign=m]{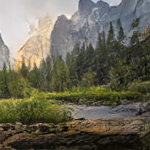}
			& \includegraphics[width=4em, valign=m]{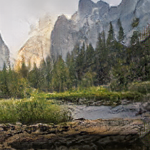}
			& \includegraphics[width=4em, valign=m]{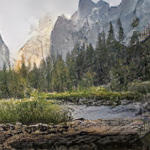}
			& \includegraphics[width=4em, valign=m]{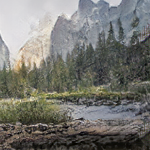}
			& \includegraphics[width=4em, valign=m]{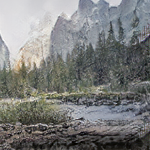}
			& \includegraphics[width=4em, valign=m]{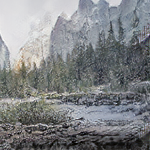}
			& \includegraphics[width=4em, valign=m]{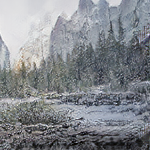}
			& \includegraphics[width=4em, valign=m]{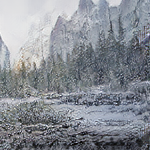}
			& \includegraphics[width=4em, valign=m]{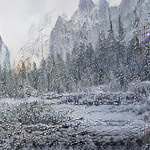}
			& \includegraphics[width=4em, valign=m]{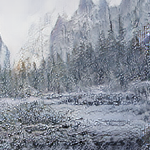}
			\\

			\multirow{1}{*}[0.25cm]{\rotatebox{90}{\tiny{SAVI2I}}}
			& \includegraphics[width=4em, valign=m]{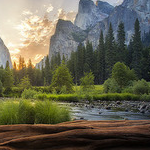}
			& ~ 
			& ~	
			& \includegraphics[width=4em, valign=m]{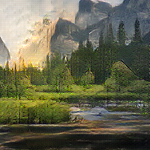}
			& \includegraphics[width=4em, valign=m]{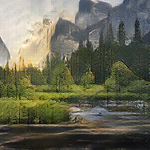}
			& \includegraphics[width=4em, valign=m]{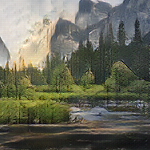}
			& \includegraphics[width=4em, valign=m]{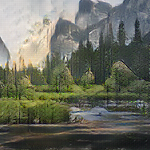}
			& \includegraphics[width=4em, valign=m]{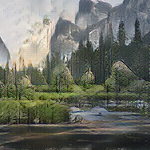}
			& \includegraphics[width=4em, valign=m]{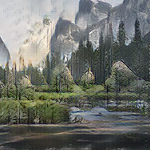}
			& \includegraphics[width=4em, valign=m]{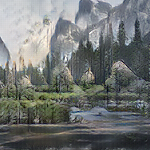}
			& \includegraphics[width=4em, valign=m]{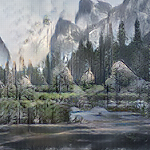}
			& \includegraphics[width=4em, valign=m]{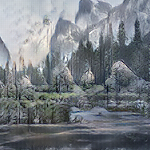}
			& \includegraphics[width=4em, valign=m]{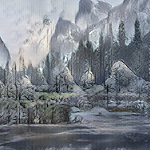}
			& \includegraphics[width=4em, valign=m]{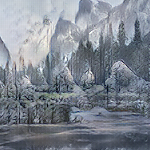}
			\\

			\multirow{1}{*}[0.35cm]{\rotatebox{90}{\tiny{MonoPix}}}
			& \includegraphics[width=4em, valign=m]{IMG_DownSample/S2W/Starganv2/Rdm/121_enhlvl_99.png}
			& ~ 
			& ~	
			& \includegraphics[width=4em, valign=m]{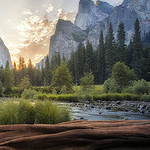}
			& \includegraphics[width=4em, valign=m]{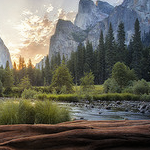}
			& \includegraphics[width=4em, valign=m]{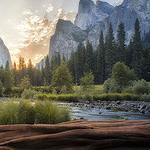}
			& \includegraphics[width=4em, valign=m]{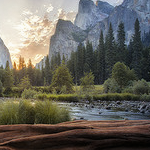}
			& \includegraphics[width=4em, valign=m]{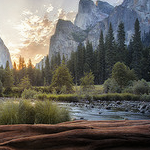}
			& \includegraphics[width=4em, valign=m]{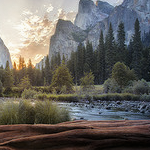}
			& \includegraphics[width=4em, valign=m]{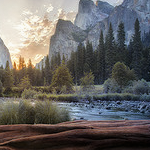}
			& \includegraphics[width=4em, valign=m]{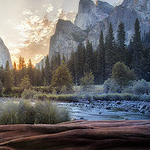}
			& \includegraphics[width=4em, valign=m]{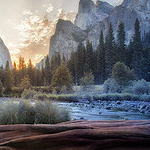}
			& \includegraphics[width=4em, valign=m]{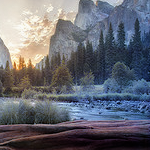}
			& \includegraphics[width=4em, valign=m]{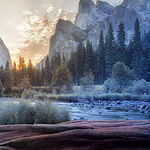}
			\\
			\multicolumn{15}{c}{Cat $\mapsto$ Dog}\\
			\multirow{1}{*}[0.4cm]{\rotatebox{90}{\tiny{StarGANv2}}}
			& \includegraphics[width=4em, valign=m]{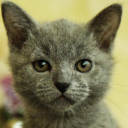}
			& ~ 
			& ~	
			& \includegraphics[width=4em, valign=m]{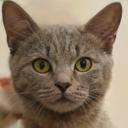}
			& \includegraphics[width=4em, valign=m]{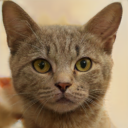}
			& \includegraphics[width=4em, valign=m]{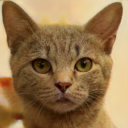}
			& \includegraphics[width=4em, valign=m]{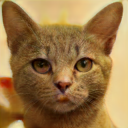}
			& \includegraphics[width=4em, valign=m]{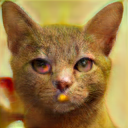}
			& \includegraphics[width=4em, valign=m]{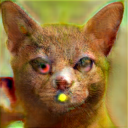}
			& \includegraphics[width=4em, valign=m]{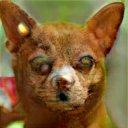}
			& \includegraphics[width=4em, valign=m]{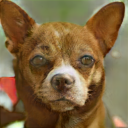}
			& \includegraphics[width=4em, valign=m]{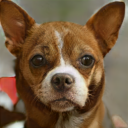}
			& \includegraphics[width=4em, valign=m]{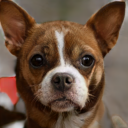}
			& \includegraphics[width=4em, valign=m]{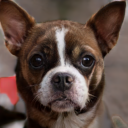}
			\\

			\multirow{1}{*}[0.4cm]{\rotatebox{90}{\tiny{Liu \etal}}}
			& \includegraphics[width=4em, valign=m]{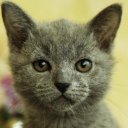}
			& ~ 
			& ~	
			& \includegraphics[width=4em, valign=m]{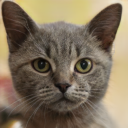}
			& \includegraphics[width=4em, valign=m]{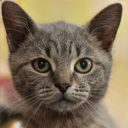}
			& \includegraphics[width=4em, valign=m]{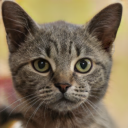}
			& \includegraphics[width=4em, valign=m]{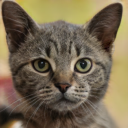}
			& \includegraphics[width=4em, valign=m]{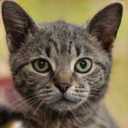}
			& \includegraphics[width=4em, valign=m]{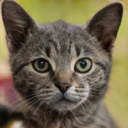}
			& \includegraphics[width=4em, valign=m]{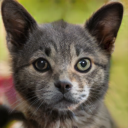}
			& \includegraphics[width=4em, valign=m]{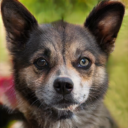}
			& \includegraphics[width=4em, valign=m]{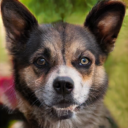}
			& \includegraphics[width=4em, valign=m]{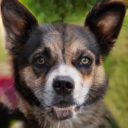}
			& \includegraphics[width=4em, valign=m]{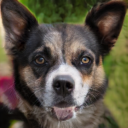}
			\\

			\multirow{1}{*}[0.25cm]{\rotatebox{90}{\tiny{SAVI2I}}}
			& \includegraphics[width=4em, valign=m]{IMG_DownSample/C2D/Starganv2/Rdm/105_enhlvl_99.png}
			& ~ 
			& ~	
			& \includegraphics[width=4em, valign=m]{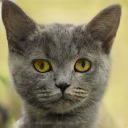}
			& \includegraphics[width=4em, valign=m]{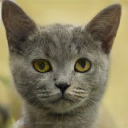}
			& \includegraphics[width=4em, valign=m]{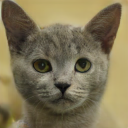}
			& \includegraphics[width=4em, valign=m]{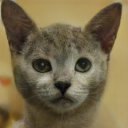}
			& \includegraphics[width=4em, valign=m]{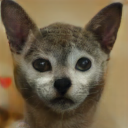}
			& \includegraphics[width=4em, valign=m]{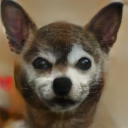}
			& \includegraphics[width=4em, valign=m]{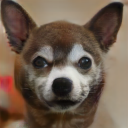}
			& \includegraphics[width=4em, valign=m]{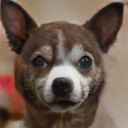}
			& \includegraphics[width=4em, valign=m]{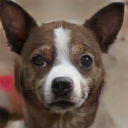}
			& \includegraphics[width=4em, valign=m]{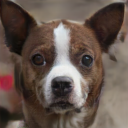}
			& \includegraphics[width=4em, valign=m]{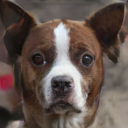}
			\\
			
			\multirow{1}{*}[0.35cm]{\rotatebox{90}{\tiny{MonoPix}}}
			& \includegraphics[width=4em, valign=m]{IMG_DownSample/C2D/Starganv2/Rdm/105_enhlvl_99.png}
			& ~ 
			& ~	
			& \includegraphics[width=4em, valign=m]{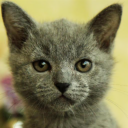}
			& \includegraphics[width=4em, valign=m]{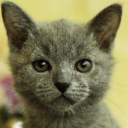}
			& \includegraphics[width=4em, valign=m]{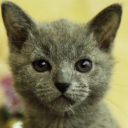}
			& \includegraphics[width=4em, valign=m]{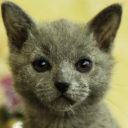}
			& \includegraphics[width=4em, valign=m]{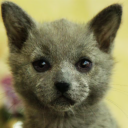}
			& \includegraphics[width=4em, valign=m]{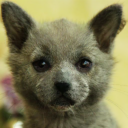}
			& \includegraphics[width=4em, valign=m]{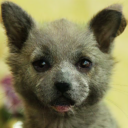}
			& \includegraphics[width=4em, valign=m]{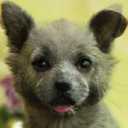}
			& \includegraphics[width=4em, valign=m]{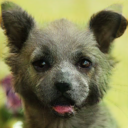}
			& \includegraphics[width=4em, valign=m]{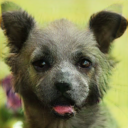}
			& \includegraphics[width=4em, valign=m]{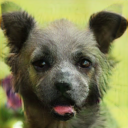}
			\\
			~ & \tikzmark{c}{}&~&~&~&~&~&~&~&~&~&~&~&~&\tikzmark{d}{} 
			\\
			~ & 
			input & 
			~ &	 
			~ &	 
			$\boldsymbol{c}=0.0$ &
			$\boldsymbol{c}=0.1$ &
			$\boldsymbol{c}=0.2$ &
			$\boldsymbol{c}=0.3$ &
			$\boldsymbol{c}=0.4$ &
			$\boldsymbol{c}=0.5$ &
			$\boldsymbol{c}=0.6$ &
			$\boldsymbol{c}=0.7$ &
			$\boldsymbol{c}=0.8$ &
			$\boldsymbol{c}=0.9$ &
			$\boldsymbol{c}=1.0$ &
		\end{tabular}\link{c}{d}}
	\caption{Visual comparisons on continuous domain translation} 
	\label{fig:DomaonTranslation}
\end{figure*}

\subsection{Implementation Details}
Following \cite{jiang2021enlightengan}, we use the U-net \cite{ronneberger2015u} structure as the generator and discriminator, and Adam as the optimizer with $\beta_1$ and $\beta_2$ set to be 0.5 and 0.9 respectively. LSGAN \cite{mao2017least} is adopted to provide the basic GAN loss. As to the hyperparameter settings, we always set $\lambda_{cyc}$ as 10, as in CycleGAN, and $\lambda_{mn}$ as 1, which is the same as the weight of $\mathcal{L}_{adv}$. $\lambda_{df}$ and $\epsilon$, along with training epochs vary in different datasets, which can be found in detail from the supplementary material. The learning rate is set to be $10^{-4}$ initially and linearly drops to zero.

\setlength{\tabcolsep}{4pt}
\def\arraystretch{1.2}
\begin{table*}[b!]
	\begin{center}
		\caption{Quantitative results on cross-domain continuous translation}
		\label{table:DoaminTranslation}
		\scalebox{0.75}{
			\begin{tabular}{lllllll}
				\toprule
				\mmc{\multirow{2}{*}{Methods}}&\multicolumn{3}{l|}{Summer $\mapsto$ Winter}&\multicolumn{3}{l}{Winter $\mapsto$ Summer}\\ \cline{2-7}
				\mmc{~} & \mmc{AL$\uparrow$/Rg$\uparrow$} & \mmc{RL$\uparrow$/Sm$\downarrow$} & \mmc{ACC$\uparrow$/FID$\downarrow$} 
				& \mmc{AL$\uparrow$/Rg$\uparrow$} & \mmc{RL$\uparrow$/Sm$\downarrow$} & \mmcn{ACC$\uparrow$/FID$\downarrow$}\\ \cline{1-7}
				
				\mmc{CycleGAN-DNI \cite{wang2019deep}} & \mmc{0.969/0.286} & \mmc{0.937/0.067} & \mmc{0.766/35.0} & \mmc{0.979/0.279} & \mmc{0.911/0.079} & \mmcn{0.834/42.8} \\ 
				\cline{1-7}
				\mmc{StarGANv2-Rdm \cite{choi2020stargan}} & \mmc{0.805/0.134} & \mmc{0.992/0.021} & \mmc{0.616/46.1} & \mmc{0.741/0.107} & \mmc{0.996/0.018} & \mmcn{0.771/50.1} \\ 
				\cline{1-7}
				\mmc{StarGANv2-Cent \cite{choi2020stargan}} & \mmc{0.719/0.100} & \mmc{0.985/0.016} & \mmc{0.650/49.3} & \mmc{0.417/0.055} & \mmc{0.996/0.010} & \mmcn{0.720/52.8}\\ 
				\cline{1-7}
				
				\mmc{SAVI2I-Rdm \cite{mao2022continuous}} & \mmc{0.959/0.185} & \mmc{0.994/0.013} & \mmc{0.680/44.5} & \mmc{0.939/0.168} & \mmc{0.995/0.013} & \mmcn{0.706/47.6} \\ 
				\cline{1-7}
				\mmc{SAVI2I-Cent \cite{mao2022continuous}} & \mmc{0.934/0.145} & \mmc{0.988/0.010} & \mmc{0.718/47.5} & \mmc{0.873/0.111} & \mmc{0.993/0.009} & \mmcn{0.721/49.2} \\ 
				\cline{1-7}
				
				\mmc{\name (Ours)} & \mmc{0.939/0.154} & \mmc{0.940/0.015} & \mmc{0.950/38.1} & \mmc{0.960/0.145} & \mmc{0.964/0.019} & \mmcn{0.949/47.3} \\ 
				\cline{1-7}

				\hline\hline
				\mmc{\multirow{2}{*}{Methods}}&\multicolumn{3}{l|}{Cat $\mapsto$ Dog}&\multicolumn{3}{l}{Dog $\mapsto$ Cat}\\ \cline{2-7}
				\mmc{~} & \mmc{AL$\uparrow$/Rg$\uparrow$} & \mmc{RL$\uparrow$/Sm$\downarrow$} & \mmc{ACC$\uparrow$/FID$\downarrow$} 
				& \mmc{AL$\uparrow$/Rg$\uparrow$} & \mmc{RL$\uparrow$/Sm$\downarrow$} & \mmcn{ACC$\uparrow$/FID$\downarrow$}\\ \cline{1-7}

				\mmc{StarGANv2-Rdm \cite{choi2020stargan}} & \mmc{0.872/0.279} & \mmc{0.979/0.299} & \mmc{0.924/33.9}  & \mmc{0.840/0.293} & \mmc{0.974/0.295} & \mmcn{0.980/25.7} \\ 
				\cline{1-7}
				\mmc{StarGANv2-Cent \cite{choi2020stargan}} & \mmc{0.738/0.242} & \mmc{0.980/0.284} & \mmc{0.906/50.2} & \mmc{0.742/0.263} & \mmc{0.974/0.282} & \mmcn{0.991/38.0} \\ 
				\cline{1-7}

				\mmc{Liu \etal -Rdm \cite{liu2021smoothing}} & \mmc{0.930/0.281} & \mmc{0.967/0.174} & \mmc{0.956/37.1}  & \mmc{0.933/0.313} & \mmc{0.956/0.216} & \mmcn{0.997/27.5} \\ 
				\cline{1-7}
				
				\mmc{SAVI2I-Rdm \cite{mao2022continuous}} & \mmc{0.973/0.294} & \mmc{0.990/0.142} & \mmc{0.942/28.6}  &\mmc{0.968/0.292} & \mmc{0.992/0.138} & \mmcn{0.982/24.2} \\ 
				\cline{1-7}
				\mmc{SAVI2I-Cent \cite{mao2022continuous}} & \mmc{0.955/0.235} & \mmc{0.988/0.141} & \mmc{0.976/53.6} & \mmc{0.957/0.251} & \mmc{0.995/0.132} & \mmcn{0.997/31.4}\\ 
				\cline{1-7}		
				
				\mmc{\name (Ours)} & \mmc{0.986/0.398} & \mmc{0.990/0.050} & \mmc{0.970/41.1} & \mmc{0.983/0.407} & \mmc{0.984/0.076} & \mmcn{0.997/14.4} \\ 
				
				\bottomrule
			\end{tabular}
		}
	\end{center}
\end{table*}

\subsection{Results on Continuous Cross-Domain Translation}
\label{sec:res_domaintrans}
We compare \name with deep network interpolation method CycleGAN-DNI \cite{wang2019deep}, style interpolation method StarGANv2 \cite{choi2020stargan}, and recently proposed continuous cross-domain translation model Liu \etal \cite{liu2021smoothing} and SAVI2I \cite{mao2022continuous}. For style-guided methods, we randomly select their reference images from the target domain to provide the style information and report the average scores in 5 experiments (we call this variant ``-Rdm''), and calculate the average target style information (style centroid) so that they can generate deterministic interpolations with $O(1)$ complexity (we call this variant ``-Cent''). 

The visual and quantitative results can be found in Figure \ref{fig:DomaonTranslation} and Table \ref{table:DoaminTranslation} respectively. We first notice that two-stage interpolation method CycleGAN-DNI and our contrastive modulation approach \name show a better fidelity on the input images, while style-guided solutions bear an obvious distortion on the original textures and attributes. 
Though CycleGAN-DNI exhibits the highest control range in Table \ref{table:DoaminTranslation}, the intermediate smoothness is significantly worse than the rest candidates. 
We also find that \name provides a competitive absolute linearity, but the modifications between adjacent translation intensities are not strongly guaranteed to be evenly distributed, as in Yosemite scenario. This is partially related to the non-linearity of domain discriminator, but is not a key practical problem since we can execute a finer exploration in ``intensity of interests'', which we illustrate more in the supplementary material. As to the realness and quality, \name also performs favorably against other state of the arts, and strongly demonstrates the benefits of introducing a contrastive modulation scheme that helps to explore domain-specific features. Nevertheless, we note that evaluating FID on the whole trajectory, or on a specific intensity may produce different results for \namewo, as we do not assume all input images share the same optimal translation intensity. More discussions on the qualitative evaluations can be found in the supplementary material. It is also noteworthy to mention that interpolating towards the style centroid leads to a smoother and deterministic translation (variant ``-Cent''), yet taking such a trajectory can cause duplicated or unrealistic patterns for style-guided approaches.

\subsection{Applications in Low-Level Image Processing}


\begin{figure*}[b!]
\setlength{\tabcolsep}{4pt}
\def\arraystretch{1.2}
\begin{minipage}[t]{\textwidth}
	\begin{minipage}[t]{0.55\textwidth}
		\centering
		\makeatletter\def\@captype{table}\makeatother\caption{Quantitative results on LOL test-set. ELGAN stands for EnlightenGAN}
		\label{table:res_LOL}
		\scalebox{0.7}{
			\begin{tabular}{lllll}
				\hline
				\mmc{Method}&\mmc{Type}&\mmc{PSNR$\uparrow$}&\mmc{SSIM$\uparrow$}&\mmcn{LPIPS$\downarrow$} \\ 
				\hline
				\mmc{TBEFN \cite{lu2020tbefn}}&\mmc{SL-O2O}&\mmc{17.35}&\mmc{0.777}&\mmcn{0.210} \\ 
				\cline{1-5}
				\mmc{KinD++ \cite{zhang2021beyond}}&\mmc{SL-Ref}&\mmc{21.80}&\mmc{0.829}&\mmcn{0.158} \\ 
				\mmc{KinD++ \cite{zhang2021beyond}}&\mmc{SL-O2O}&\mmc{17.75}&\mmc{0.758}&\mmcn{0.198} \\ 
				\cline{1-5}
				\mmc{ELGAN \cite{jiang2021enlightengan}}&\mmc{US-O2O}&\mmc{19.01}&\mmc{0.709}&\mmcn{0.274} \\ 
				\cline{1-5}
				\mmc{Ours-EI}&\mmc{US-Ref}&\mmc{21.65}&\mmc{0.731}&\mmcn{0.273} \\ 
				\mmc{Ours-TS}&\mmc{US-Ref}&\mmc{21.55}&\mmc{0.731}&\mmcn{0.273} \\ 
				\mmc{Ours-LP}&\mmc{Few Shot}&\mmc{19.75}&\mmc{0.725}&\mmcn{0.272} \\
				\hline
			\end{tabular}
		}
	\end{minipage}
	\begin{minipage}[t]{0.45\textwidth}
		\centering
		\makeatletter\def\@captype{table}\makeatother\caption{Quantitative results on SIDD natural noise generation}
		\label{table:res_SIDD}
		\scalebox{0.8}{
			\begin{tabular}{lll}
				\hline
				\mmc{Method}&\mmc{Type}&\mmcn{AKLD$\downarrow$} \\ 
				\hline
				\mmc{CBDNet \cite{guo2019toward}}&\mmc{\multirow{4}{*}{SL}}&\mmcn{0.728} \\ 
				\mmc{ULRD \cite{brooks2019unprocessing}}&\mmc{~}&\mmcn{0.545} \\ 
				\mmc{GRDN \cite{kim2019grdn}}&\mmc{~}&\mmcn{0.443} \\ 
				\mmc{DANet \cite{yue2020dual}}&\mmc{~}&\mmcn{0.212} \\ 
				\hline
				\mmc{Ours-EI}&\mmc{\multirow{2}{*}{US}}&\mmcn{0.137} \\ 
				\mmc{Ours-TS}&\mmc{~}&\mmcn{0.139} \\ 
				\hline
			\end{tabular}
		}
	\end{minipage}
\end{minipage}
\end{figure*}

In addition to cross-domain continuous translation, we show how \namewo, along with the inference strategies can be used to facilitate low-level tasks. 

\noindent\textbf{Low-Light Image Enhancement.}
We integrate the proposed contrastive modulation approach with EnlightenGAN \cite{jiang2021enlightengan}, a previous unsupervised one-to-one mapping (US-O2O) method in this area, and compare the performance on LOL test set with several state-of-the-art methods, including supervised one-to-one mapping (SL-O2O) method TBEFN \cite{lu2020tbefn}, and reference-guided supervised (SL-Ref) approach KinD++ \cite{zhang2021beyond}. For fairness, we retrain EnlightenGAN on LOL training set, which yields an improved performance. 
We choose PSNR to guide the ternary search for this task unless specified.

\begin{figure*}[t!]
	\centering
	\resizebox{\linewidth}{!}{
		\setlength{\tabcolsep}{0.003\linewidth}
		\large
		\begin{tabular}{c c c c c c}
			\multicolumn{6}{c}{\small Low light $\mapsto$ Normal light}\\
			\includegraphics[width=6em, valign=m]{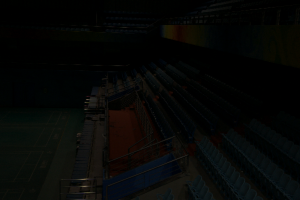}
			& \includegraphics[width=6em, valign=m]{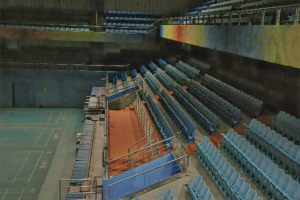}
			& \includegraphics[width=6em, valign=m]{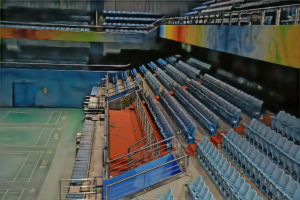}
			& \includegraphics[width=6em, valign=m]{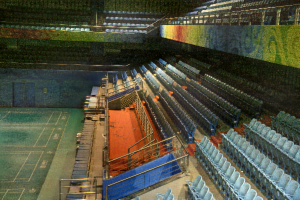}
			& \includegraphics[width=6em, valign=m]{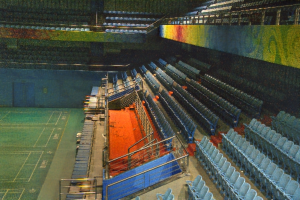}
			& \includegraphics[width=6em, valign=m]{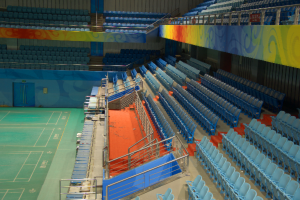}
			\\
			\small Input &
			\small TBEFN &
			\small KinD++ Ref &
			\small EnlightenGAN &
			\small MonoPix TS &
			\small Reference

			\\\multicolumn{6}{c}{\small Clean $\mapsto$ Noisy}\\
			\includegraphics[width=4em, valign=m]{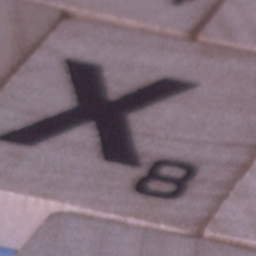}
			& \includegraphics[width=4em, valign=m]{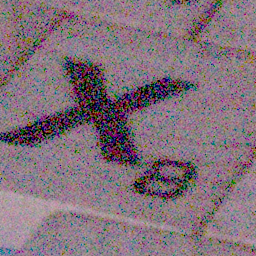}
			& \includegraphics[width=4em, valign=m]{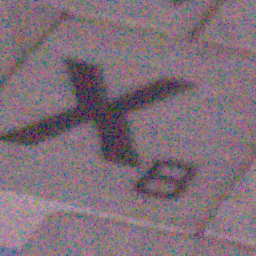}
			& \includegraphics[width=4em, valign=m]{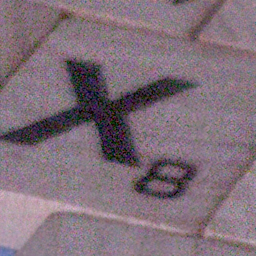}
			& \includegraphics[width=4em, valign=m]{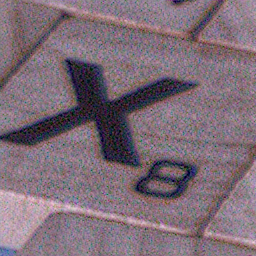}
			& \includegraphics[width=4em, valign=m]{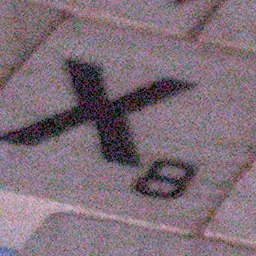}
			\\
			\small Input &
			\small CBDNet &
			\small ULRD &
			\small DANet &
			\small MonoPix TS &
			\small Reference
			
		\end{tabular}}
	\caption{Visual comparisons on the LOL low-light enhancement task and SIDD natural noise generation}
	\label{fig:res_lowlevel}
\end{figure*}

To figure out the potential of our controllable generation scheme, we first implement exhaustive inference (EI), whose results can be viewed as the upper bound. As can be found in Table \ref{table:res_LOL}, \name yields an average PSNR of 21.65, which is considerably higher than the one-to-one mapping baselines. This is because a rigid O2O mapping is not sufficient to model the diversity of data distribution. Interestingly, this result is even comparable to state-of-the-art reference-guided supervised approach KinD++, but note that \name can produce competitive results without any expert knowledge or paired supervision, and streamlines the enhancement pipeline notably. The superiority of \name is also supported from SSIM and LPIPS scores. When using a 7-times ternary search, we observe a tiny and acceptable performance drop. This evidence not only validates the monotonicity and continuity of the shaped generation manifold, but also demonstrates the effectiveness of our inference strategy. 
Next, we formulate a non-reference-guided variant by training an expert network with 10\% of paired supervision (during training the backbone conditional generator is frozen). Shown in the experiment named ``LP'', the model then degrades to an O2O mapping-based translator where the previous unsupervised contrastive training stage now serves as a pre-text task. Though not competitive as the ternary search variant, it still produces competitive results, yet with a complexity of only $O(1)$. 
Visual comparisons on this task can be found in Figure \ref{fig:res_lowlevel}.

\begin{figure*}[t!]
	\centering
	\resizebox{\linewidth}{!}{
		\setlength{\tabcolsep}{0.003\linewidth}
		\large
		\begin{tabular}{c c c c c c c | c c c c c c c}

			\includegraphics[width=4em, valign=m]{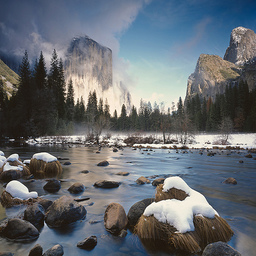}
			& +
			& \includegraphics[width=4em, valign=m]{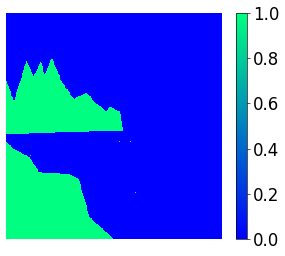}
			& =
			& \includegraphics[width=4em, valign=m]{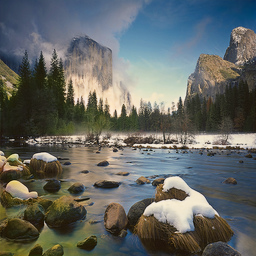}
			& ~
			& ~			
			& ~			
			& ~
			& \includegraphics[width=4em, valign=m]{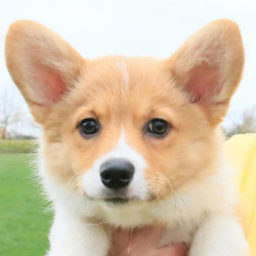}			
			& +
			& \includegraphics[width=4em, valign=m]{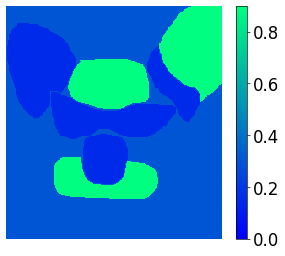}
			& =
			& \includegraphics[width=4em, valign=m]{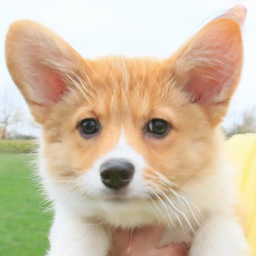}
			\\		
			\hline
			\includegraphics[width=4em, valign=m]{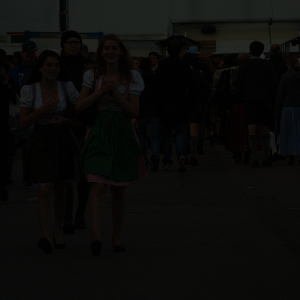}
			& +
			& \includegraphics[width=4em, valign=m]{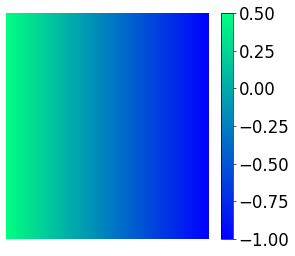}
			& =
			& \includegraphics[width=4em, valign=m]{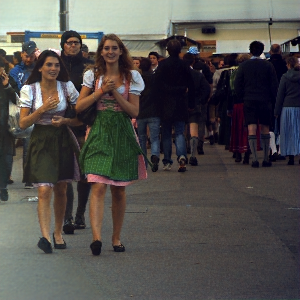}
			& ~
			& ~
			& ~
			& ~			
			& \includegraphics[width=4em, valign=m]{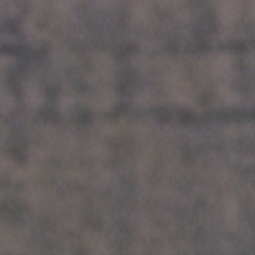}			
			& +
			& \includegraphics[width=4em, valign=m]{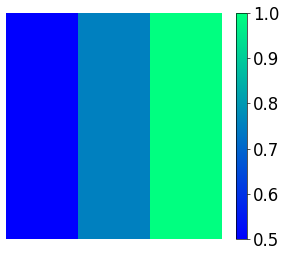}
			& =
			& \includegraphics[width=4em, valign=m]{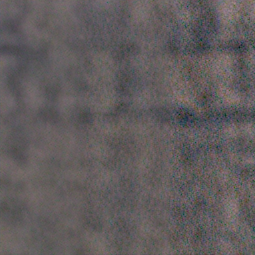}	

		\end{tabular}}
	\caption{Examples on fine-grained pixel-level control. From left to right, we provide the input image, spatial control signal, and the modulation result respectively}
	\label{fig:spatial}
\end{figure*}

\noindent\textbf{Blind Image Noise Generation.}
Different from the commonly discussed task of image denoising, noise generation takes the reverse consideration, and typically is an O2M mapping task. We show in this part the potential and benefits of introducing the proposed continuous modulation approach into this field, and compare with state-of-the-art methods CBDNet \cite{guo2019toward}, ULRD \cite{brooks2019unprocessing}, GRDN \cite{kim2019grdn}, and DANet \cite{yue2020dual}. Quantitative and visual results can be found in Table \ref{table:res_SIDD} and Figure \ref{fig:res_lowlevel} respectively, where our model (``TS'' variant is guided by AKLD) again exhibits its superiority and generalization ability. 
To the best of our knowledge, \name is the first work that proposes to continuously model natural noise in an unsupervised and spatially-controlled manner. More visual comparisons can be found in the supplementary material.

\noindent\textbf{Pixel-level Monotonic Control.}
\label{sec:oob}
In all the previous experiments, we change the overall translation strength by assigning each pixel with the same intensity. Recall that a fine-grained pixel-level control can be naturally obtained by customizing $c^{i,j}$, we here provide a few examples in Figure \ref{fig:spatial}. Note that the linearity of manifold further enables a slightly out-of-the-bound inference, especially for uni-directional tasks like illumination and noise control.

\def\arraystretch{1.3}
\begin{table*}[b!]
	\begin{center}
		\caption{Ablation studies on Yosemite dataset}
		\label{table:res_ablation}
		\scalebox{0.75}{
			\begin{tabular}{llllllllll}
				\hline
				\mmc{\multirow{2}{*}{Variant}}&\mmc{\multirow{2}{*}{$\mathcal{L}_{mono}$}}&\mmc{\multirow{2}{*}{$\mathcal{L}_{df}$}}&\mmc{\multirow{2}{*}{$\epsilon$}}&\multicolumn{3}{l|}{Summer $\rightarrow$ Winter}&\multicolumn{3}{l}{Winter $\rightarrow$ Summer}\\ \cline{5-10}
				\mmc{~} & \mmc{~} & \mmc{~} & \mmc{~} & \mmc{AL$\uparrow$/Rg$\uparrow$} & \mmc{RL$\uparrow$/Sm$\downarrow$} & \mmc{ACC$\uparrow$/FID$\downarrow$} & \mmc{AL$\uparrow$/Rg$\uparrow$} & \mmc{RL$\uparrow$/Sm$\downarrow$} & \mmcn{ACC$\uparrow$/FID$\downarrow$}\\ \cline{1-10}
				\mmc{(a)} & \mmc{~} & \mmc{~} & \mmc{\multirow{3}{*}{0.5}} & \mmc{0.143/0.000} & \mmc{0.999/0.000} & \mmc{0.733/49.9} & \mmc{0.105/0.000} & \mmc{1.000/0.000} & \mmcn{0.702/60.0} \\ 
				\cline{1-3}\cline{5-10}
				\mmc{(b)} & \mmc{$\surd$} & \mmc{~} & \mmc{~} & \mmc{0.927/0.091} & \mmc{0.914/0.008} & \mmc{0.866/41.3}  & \mmc{0.963/0.097} & \mmc{0.964/0.010} & \mmcn{0.847/50.3} \\ 
				\cline{1-3}\cline{5-10}
				\mmc{(c)} & \mmc{$\surd$} & \mmc{$\surd$} & \mmc{~} & \mmc{0.939/0.154} & \mmc{0.940/0.015} & \mmc{0.950/38.1} & \mmc{0.960/0.145} & \mmc{0.964/0.019} & \mmcn{0.949/47.3} \\ \cline{1-3}\cline{4-10}
				\mmc{(d)} & \mmc{$\surd$} & \mmc{$\surd$} & \mmc{0.0} & \mmc{0.415/0.000} & \mmc{1.000/0.000} & \mmc{0.735/50.0} & \mmc{0.250/0.000} & \mmc{1.000/0.000} & \mmcn{0.712/59.5} \\ \hline
				\mmc{(e)} & \mmc{$\surd$} & \mmc{$\surd$} & \mmc{1.0} & \mmc{0.771/0.569} & \mmc{0.861/0.210} & \mmc{0.996/39.7} & \mmc{0.774/0.571} & \mmc{0.873/0.204} & \mmcn{0.994/48.7} \\
				\hline	
			\end{tabular}
		}
	\end{center}
\end{table*}

\section{Discussion}
\subsection{Ablation Studies}
\label{sec:ablation}
To further examine the impact of the introduced method, we conduct ablation studies on Yosemite summer-winter translation with two key components that construct our approach: the monotonicity loss (carried along with the contrastive training scheme) and domain fidelity loss. Shown in Table \ref{table:res_ablation} (a) and Figure \ref{fig:ablation}, the variant with none of these losses degrades to an O2O mapping, as we have expected. Note that the AL score is near zero since Pearson correlation coefficient can be negative. When we further add the monotonicity loss, the model succeeds in catching confidence differences and produces a coarse yet continuous translation. 
Our full model, as shown in variant (c), yields a better fidelity and wider control range. We find the translation quality is also improved under this setting, which is because fidelity loss also strengthens the translation monotonicity and encourages to explore more domain-specific features.

\subsection{Sensitivity Analysis}

\begin{figure*}[t]
\begin{minipage}[t]{\textwidth}
	\begin{minipage}[b]{0.65\textwidth}
		\centering
		\resizebox{\linewidth}{!}{
			\setlength{\tabcolsep}{0.003\linewidth}
			\tiny
			\begin{tabular}{c c | c c c c c c}
				\multirow{1}{*}[1.2cm]{\rotatebox{90}{\emph{w/o} $\mathcal{L}_{mono}$}}
				& \begin{overpic}[width=6em, valign=m]{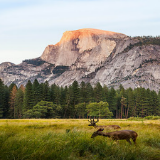}
					\put(0,90){\tiny\color{blue}{~}}
				\end{overpic}
				& \begin{overpic}[width=6em, valign=m]{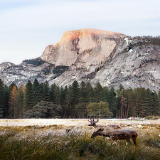}
					\put(0,90){\tiny\color{blue}{LPIPS: 0.164}}
				\end{overpic}
				& \begin{overpic}[width=6em, valign=m]{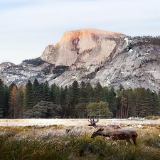}
					\put(0,90){\tiny\color{blue}{LPIPS: 0.164}}
				\end{overpic}
				& \begin{overpic}[width=6em, valign=m]{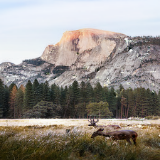}
					\put(0,90){\tiny\color{blue}{LPIPS: 0.164}}
				\end{overpic}
				& \begin{overpic}[width=6em, valign=m]{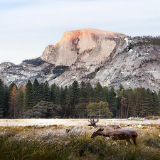}
					\put(0,90){\tiny\color{blue}{LPIPS: 0.164}}
				\end{overpic}	
				& \begin{overpic}[width=6em, valign=m]{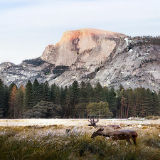}
					\put(0,90){\tiny\color{blue}{LPIPS: 0.164}}
				\end{overpic}
				& \begin{overpic}[width=6em, valign=m]{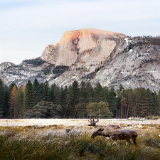}
					\put(0,90){\tiny\color{blue}{LPIPS: 0.164}}
				\end{overpic}					
				\\
				
				\multirow{1}{*}[0.9cm]{\rotatebox{90}{\emph{w/o} $\mathcal{L}_{df}$}}
				& \begin{overpic}[width=6em, valign=m]{IMG_DownSample/S2W/MonoPix/Ablation_NoMono_293/293_enhlvl_99.png}
					\put(0,90){\tiny\color{blue}{~}}
				\end{overpic}
				& \begin{overpic}[width=6em, valign=m]{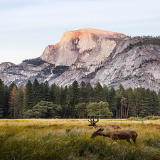}
					\put(0,90){\tiny\color{blue}{LPIPS: 0.031}}
				\end{overpic}
				& \begin{overpic}[width=6em, valign=m]{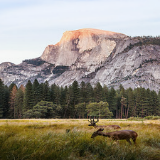}
					\put(0,90){\tiny\color{blue}{LPIPS: 0.038}}
				\end{overpic}
				& \begin{overpic}[width=6em, valign=m]{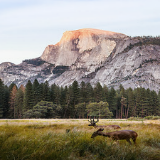}
					\put(0,90){\tiny\color{blue}{LPIPS: 0.057}}
				\end{overpic}
				& \begin{overpic}[width=6em, valign=m]{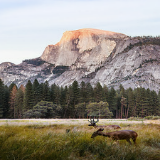}
					\put(0,90){\tiny\color{blue}{LPIPS: 0.083}}
				\end{overpic}	
				& \begin{overpic}[width=6em, valign=m]{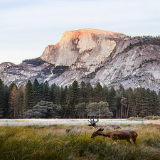}
					\put(0,90){\tiny\color{blue}{LPIPS: 0.114}}
				\end{overpic}
				& \begin{overpic}[width=6em, valign=m]{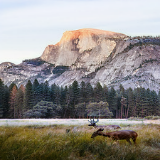}
					\put(0,90){\tiny\color{blue}{LPIPS: 0.148}}
				\end{overpic}						
				\\

				\multirow{1}{*}[1.1cm]{\rotatebox{90}{Full Model}}
				& \begin{overpic}[width=6em, valign=m]{IMG_DownSample/S2W/MonoPix/Ablation_NoMono_293/293_enhlvl_99.png}
					\put(0,90){\tiny\color{blue}{~}}
				\end{overpic}
				& \begin{overpic}[width=6em, valign=m]{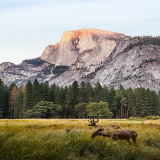}
					\put(0,90){\tiny\color{blue}{LPIPS: 0.021}}
				\end{overpic}
				& \begin{overpic}[width=6em, valign=m]{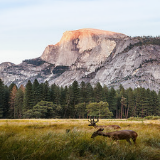}
					\put(0,90){\tiny\color{blue}{LPIPS: 0.031}}
				\end{overpic}
				& \begin{overpic}[width=6em, valign=m]{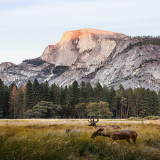}
					\put(0,90){\tiny\color{blue}{LPIPS: 0.067}}
				\end{overpic}
				& \begin{overpic}[width=6em, valign=m]{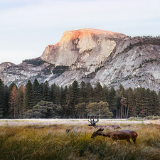}
					\put(0,90){\tiny\color{blue}{LPIPS: 0.123}}
				\end{overpic}	
				& \begin{overpic}[width=6em, valign=m]{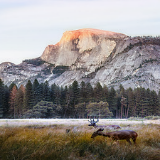}
					\put(0,90){\tiny\color{blue}{LPIPS: 0.170}}
				\end{overpic}
				& \begin{overpic}[width=6em, valign=m]{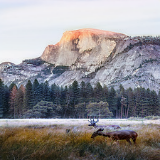}
					\put(0,90){\tiny\color{blue}{LPIPS: 0.209}}
				\end{overpic}					
				\\

				\multirow{1}{*}[0.8cm]{\rotatebox{90}{$\epsilon=1$}}
				& \begin{overpic}[width=6em, valign=m]{IMG_DownSample/S2W/MonoPix/Ablation_NoMono_293/293_enhlvl_99.png}
					\put(0,90){\tiny\color{red}{~}}
				\end{overpic}
				& \begin{overpic}[width=6em, valign=m]{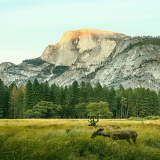}
					\put(0,90){\tiny\color{blue}{LPIPS: 0.076}}
				\end{overpic}
				& \begin{overpic}[width=6em, valign=m]{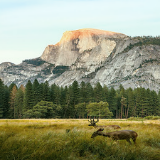}
					\put(0,90){\tiny\color{blue}{LPIPS: 0.048}}
				\end{overpic}
				& \begin{overpic}[width=6em, valign=m]{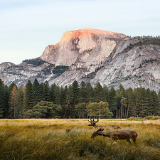}
					\put(0,90){\tiny\color{blue}{LPIPS: 0.038}}
				\end{overpic}
				& \begin{overpic}[width=6em, valign=m]{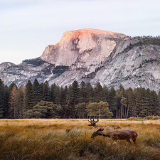}
					\put(0,90){\tiny\color{blue}{LPIPS: 0.091}}
				\end{overpic}	
				& \begin{overpic}[width=6em, valign=m]{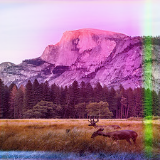}
					\put(0,90){\tiny\color{blue}{LPIPS: 0.355}}
				\end{overpic}
				& \begin{overpic}[width=6em, valign=m]{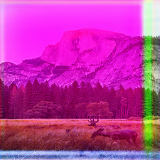}
					\put(0,90){\tiny\color{blue}{LPIPS: 0.678}}
				\end{overpic}	
				\\

				~ & \tikzmark{c}{}&~&~&~&~&~&\tikzmark{d}{} 
				\\
				~ &
				input &
				$\boldsymbol{c}=0.0$ &
				$\boldsymbol{c}=0.2$ &
				$\boldsymbol{c}=0.4$ &
				$\boldsymbol{c}=0.6$ &
				$\boldsymbol{c}=0.8$ &
				$\boldsymbol{c}=1.0$		
			\end{tabular}\link{c}{d}}
		
	\caption{Visual comparisons on the ablation study and sensitivity analysis. We label for each translated image the LPIPS score against the original input to provide a better comparison}
		\label{fig:ablation}	
	\end{minipage}
	\begin{minipage}[b]{0.3\textwidth}
		\centering
		\resizebox{\linewidth}{!}{
			\setlength{\tabcolsep}{0.003\linewidth}
			\begin{tabular}{c c c}
			\includegraphics[width=3.5em, valign=m]{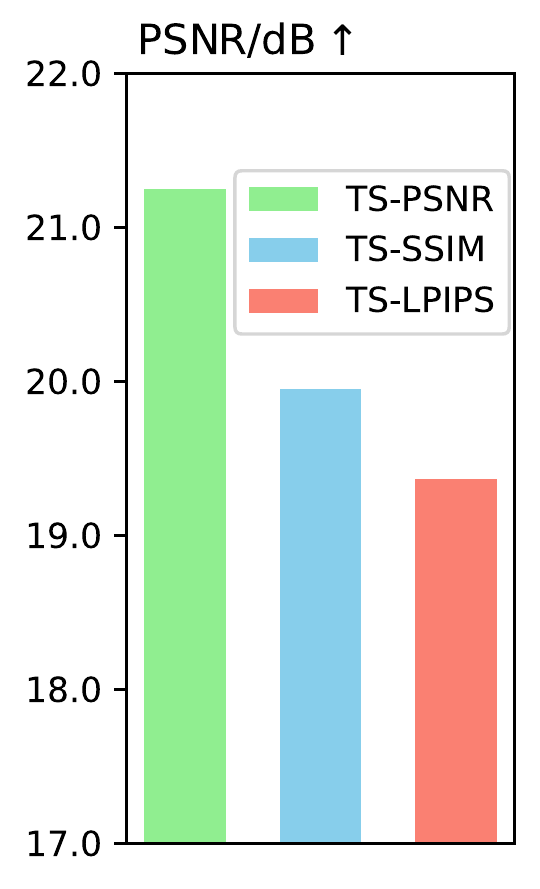} \hfil \hfil &
			\includegraphics[width=3.5em, valign=m]{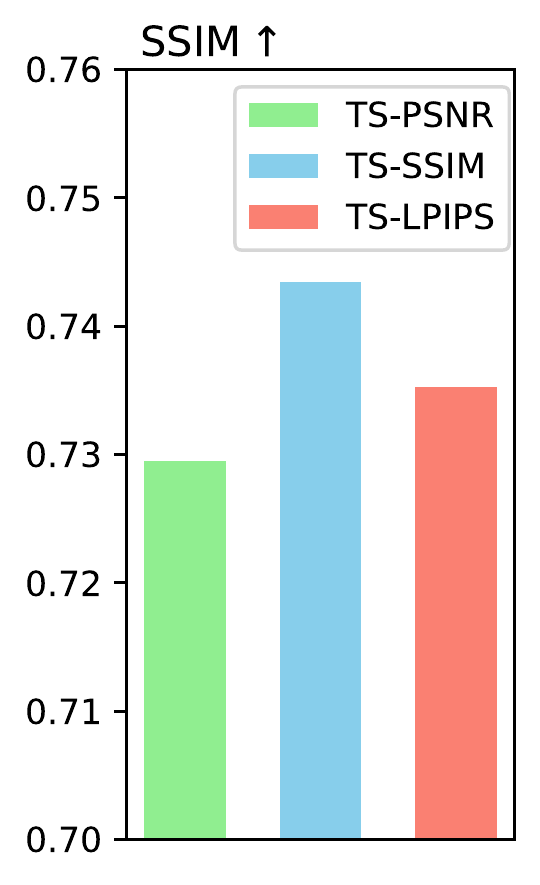} &
			\includegraphics[width=3.5em, valign=m]{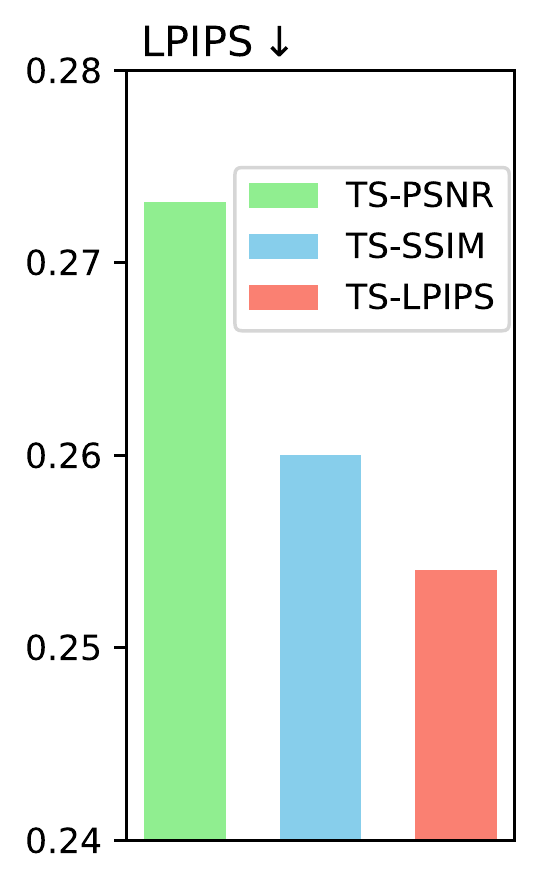} \\
			\multicolumn{3}{c}{\includegraphics[width=12em, valign=m]{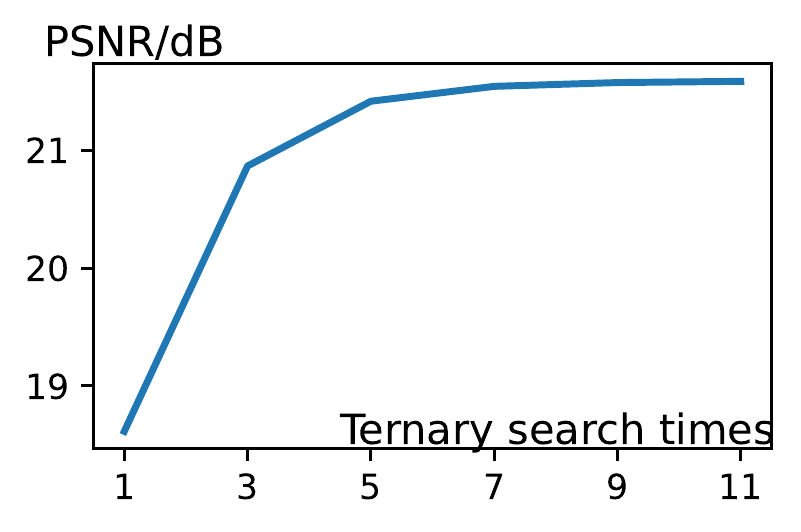}} \\
			\end{tabular}}
		\caption{Sensitivity analysis on the choice of ternary search criterion (top) and impact of ternary search complexity (bottom)}
		\label{fig:sensitivity}
	\end{minipage}
\end{minipage}
\end{figure*}

We provide the sensitivity analysis on the hyperparameter $\epsilon$ in $\mathcal{L}_{mono}$ and $\mathcal{L}_{df}$, robustness of guiding criterion in ternary search, and the impact of inference complexity in this part. For better comparisons, we label the variants with varied $\epsilon$ consecutively following Table \ref{table:res_ablation}(c). The results in experiment (d) illustrate that merely asking the discriminator to produce a higher confidence level is insufficient and easily collapses to a trivial solution. In contrast, if we set $\epsilon$ to 1.0, as in experiment (e), the whole loss function is then largely dominated by $\mathcal{L}_{mono}$ and $\mathcal{L}_{df}$, which ultimately leads to an unpleasant translation. Here we note that variant (e) achieves a better FID score than (d), as the latter suffers from a severe mode collapse issue; however considering the unrealistic translation results, setting a high $\epsilon$ inevitably introduces a quality degradation for our full model (c). 
Concluded from these experiments, hyperparameter $\epsilon$ provides a balance for control intensity and visual quality, and is suggested to increase from a small value. Typically, setting $\epsilon=0.5$ can yield acceptable translations for most of the examined tasks. Figure \ref{fig:sensitivity} shows that \name can always produce pleasant results and correspondingly achieves the best score when guided with a specific criterion. Though not surprising, these results also provide a strong evidence on the feasibility and robustness of our proposals. We have also tested the impact of $N$ which directly controls the inference complexity, where a 5 to 7 times of ternary search is found sufficient to provide competitive results.

\section{Conclusions}

We investigate in this paper an interesting yet important task - unsupervised continuous pixel-level modulation, and present \namewo, the first successful solution by devising a novel contrastive modulation framework and the corresponding constraints. \name can be deployed with selective inference complexity, and achieves a state-of-the-art performance on both the translation continuity and quality. The proposed contrastive modulation approach also provides a new alternative to paired or one-to-one mapping pipelines in low-level visions.

We also note that there are still several challenges under the current MonoPix framework, such as pixel-level multi-modal and multi-attribute control, absolute-relative linearity balance, better constraints and evaluation metrics etc. We hope the proposed method could attract more research interest in continuous pixel-level modulation, and help to shape our view on non-O2O mappings.

\noindent\textbf{Acknowledgment.} 
This work was supported in part by the National Natural Science Foundation of China under Grants 61731002 and 62071425, in part by the Zhejiang Key Research and Development Plan under Grants 2019C01002 and 2019C03131, in part by Huawei Cooperation Project, and in part by the Zhejiang Provincial Natural Science Foundation of China under Grant LY20F010016.

\bibliographystyle{splncs04}
\bibliography{monopix}

\clearpage
\appendix

\begin{center}
{\Large \bf Supplementary Material}
\end{center}

\section{Ternary Search Implementations}

We mention in Section \ref{sec:ternarysearch} of the main paper the strategy of ternary search (TS). Now we detail how it is implemented. 
Let $f(c)$ be the aesthetic score function where we tacitly assume it is concave, to find the maximum value of $f(c)$, ternary search iteratively updates two boundary points $c_{low}$ and $c_{high}$ with the time complexity of $O(\log N)$, as shown in Algorithm 1. Note that for monotonically increasing aesthetic functions, ternary search also holds as the left boundary $c_{low}$ will always yield a lower score than the right boundary point, thus resulting in a monotonic approximation towards the global maxima.

\vspace{-1cm}
\begin{figure}[h]
	\centering
	\renewcommand{\algorithmiccomment}{\textbf{// }}
	\renewcommand{\algorithmicrequire}{\textbf{Input:}}
	\renewcommand{\algorithmicensure}{\textbf{Output:}}
	\scalebox{0.75}{
	\begin{minipage}{\linewidth}
	\begin{algorithm}[H]
		\caption{Ternary Search for MonoPix}
		\begin{algorithmic}[1]
			\REQUIRE Concave or monotonically increasing aesthetic function $f(c)$, search range $[c_{left}, c_{right}]$, iteration times $N$
			\ENSURE The best enhance intensity $c_\star$
			\STATE set boundary point $c_{low} \leftarrow c_{left}$, $c_{high} \leftarrow c_{right}$			
			\FOR{epoch=1:$N$}
			\STATE mid point $c_1 \leftarrow c_{low} + (c_{high}-c_{low})/3$			
			\STATE mid point $c_2 \leftarrow c_{high} - (c_{high}-c_{low})/3$
			\IF{$f(c_1) \leq f(c_2)$}
			\STATE $c_{low} \leftarrow c_1$
			\ELSE
			\STATE $c_{high} \leftarrow c_2$
			\ENDIF
			\ENDFOR
			\STATE $c_\star \leftarrow (c_{low} + c_{high})/2$
		\end{algorithmic}
	\end{algorithm}
	\end{minipage}}
\end{figure}

\section{Network Structure and Training Details}

\subsection{Structures and Parameters}
As shown in Table \ref{table:structure}, we use the U-net \cite{ronneberger2015u} structure as the backbone for our generator, which consists of four sets of up and down-samplings. Following \cite{jiang2021enlightengan}, we replace the original deconvolution with bilinear samplings to better mitigate the checkerboard artifacts. Note that MonoPix has an extra pixel-level control signal, so it only modifies the first convolution layer with marginal extra computations. The discriminator has three stride convolution layers, and is borrowed from CycleGAN \cite{zhu2017unpaired}. Overall, the model has 8.6M parameters in generator, and 2.8M in discriminator, which is quite efficient compared with StarGANv2 \cite{choi2020stargan} (33.9M and 20.8M), CycleGAN \cite{zhu2017unpaired} (11.4M and 2.8M), and SAVI2I \cite{mao2022continuous} (6.9M not including attribute encoder, and 26.6M).

By default, we set normalization layers in MonoPix as identity mappings, which can be further adjusted for different tasks. We note that it is important to introduce certain non-linearity before instance normalization, as we have discussed in Section \ref{sec:InsNorm} of the main paper. For LOL low-light enhancement task, we integrate our contrastive modulation scheme into the authors' code instead.

\begin{table*}[t!]
	\setlength{\abovecaptionskip}{-0cm}
	\setlength{\belowcaptionskip}{-0.cm}
	\begin{center}
		\caption{The detailed structure of our generator and discriminator. We represent the parameters of a convolution layer by (input channels, output channels, kernel size, and stride). $\texttt{C}$, $\texttt{CLN}$, $\texttt{MP}$, $\texttt{UP}$, $\texttt{Cat}$ denotes convolution, convolution with leaky ReLU followed by a normalization layer, max pooling, bilinear upsampling, and concatenation respectively}
		\label{table:structure}
		\scalebox{0.75}{
		\begin{tabular}{l | l}
			\toprule
			\multicolumn{2}{c}{Generator} \\ \hline
			layer & operations \\ \hline
			1 & [\texttt{CLN}(3+1, 32, 3, 1), \texttt{CLN}(32, 32, 3, 1), \texttt{MP}(2)]\\ \hline
			2-4 & [\texttt{CLN}($C$, $2C$, 3, 1), \texttt{CLN}($2C$, $2C$, 3, 1), \texttt{MP}(2)] $\times 3 $, from $C=32$ to $128$ \\ \hline
			5 & [\texttt{CLN}(256, 512, 3, 1), \texttt{CLN}(512, 512, 3, 1)]\\ \hline
			6-9 & [\texttt{UP}(2), \texttt{C}($C$, $C$/2, 3, 1), \texttt{Cat}, \texttt{CLN}($C$, $C$/2, 3, 1), \texttt{CLN}($C$/2, $C$/2, 3, 1)] $\times 4 $, from $C=512$ to $64$ \\ \hline
			10 & \texttt{C}(32, 3, 1, 1) \\ \hline \hline
			\multicolumn{2}{c}{Discriminator} \\ \hline 
			1 & \texttt{CL}[3, 64, 4, 2] \\ \hline
			2-3 & [\texttt{CNL}($C$, $2C$, 4, 2)] $\times 2 $, from $C=64$ to $128$ \\ \hline
			4 & \texttt{CL}(256, 512, 4, 1) \\ \hline
			5 & \texttt{C}(512, 1, 4, 1) \\ \hline
			
		\end{tabular}}
	\end{center}
\end{table*}

\subsection{Training Details}

We use the official training/testing split for all datasets. The detailed implementations are described as follows:

\noindent\textbf{Yosemite Summer-Winter Translation.} The training set contains 1231 summer photos and 962 winter photos. Following the author's implementation, we resize these images to 256, and implement random horizontal flip as the augmentation. We set $\lambda_{df}$ as 0.25, and the margin $\epsilon$ as 0.5. The whole model is trained with a batch size of 8 (4$\times$2 since we generate two $\boldsymbol{c}$ for each patch) for 300 epochs, where the learning rate starts to drop linearly from epoch 200.

\noindent\textbf{AFHQ Cat-Dog Translation.} We train MonoPix with 5153 cat images and 4739 dog images. $\lambda_{df}$ is set as 0.5. The rest settings are exactly the same as in Yosemite summer-winter translation, except for the training epochs which is set to be 200. The learning rate drops from epoch 100.

\noindent\textbf{LOL Low-Light Enhancement.} We follow EnlightenGAN \cite{jiang2021enlightengan} to use the 485 LOL dark images as the source domain, and 1016 normal images as the target domain. Note that this task is concentrated on enhancing dark images only, so we train a unidirectional generator and discriminator and do not use the domain fidelity loss. The margin $\epsilon$ is set to be 0.33. For MonoPix, we train with a batch size of 32 (16$\times 2$) while for EnlightenGAN we train with batch size of 16 (so that the total iterations are the same). Both are trained for 200 epochs.

\noindent\textbf{SIDD Noise Generation.} We follow the setting in DANet \cite{yue2020dual} to process the dataset. Differently, we train MonoPix in an unsupervised and unidirectional manner, with a batch size of 16 (8$\times 2$) and patch size 128 for 200 epochs.

\section{More Experimental Analysis}

\def\arraystretch{1.3}
\begin{table*}[t!]
	\begin{center}
		\caption{More experimental results. Variant (a) is the model adopted in the main paper}
		\label{table:MoreResults}
		\scalebox{0.75}{
			\begin{tabular}{llllllll}
				\hline
				\mmc{\multirow{2}{*}{Variant}}& \mmc{\multirow{2}{*}{Description}} & \multicolumn{3}{l|}{Summer $\rightarrow$ Winter} & \multicolumn{3}{l}{Winter $\rightarrow$ Summer}\\ \cline{3-8}
				\mmc{~} & \mmc{~} & \mmc{AL$\uparrow$/Rg$\uparrow$} & \mmc{RL$\uparrow$/Sm$\downarrow$} & \mmc{ACC$\uparrow$/FID$\downarrow$} & \mmc{AL$\uparrow$/Rg$\uparrow$} & \mmc{RL$\uparrow$/Sm$\downarrow$} & \mmcn{ACC$\uparrow$/FID$\downarrow$}\\ \hline
				\mmc{(a)} & \mmc{MonoPix} & \mmc{0.939/0.154} & \mmc{0.940/0.015} & \mmc{0.950/38.1} & \mmc{0.960/0.145} & \mmc{0.964/0.019} & \mmcn{0.949/47.3} \\ \hline
				\mmc{(b)} & \mmc{$w/o \ \mathcal{L}_{df}$} & \mmc{0.927/0.091} & \mmc{0.914/0.008} & \mmc{0.866/41.3}  & \mmc{0.963/0.097} & \mmc{0.964/0.010} & \mmcn{0.847/50.3} \\ \hline
				\mmc{(c)} & \mmc{$\mathcal{L}_{df} \rightarrow L1$} & \mmc{0.896/0.111} & \mmc{0.916/0.010} & \mmc{0.893/39.8} & \mmc{0.909/0.128} & \mmc{0.930/0.016} & \mmcn{0.887/47.7} \\ \hline
				\mmc{(d)} & \mmc{Add IN} & \mmc{0.975/0.148} & \mmc{0.972/0.014} & \mmc{0.917/38.6} & \mmc{0.981/0.120} & \mmc{0.972/0.008} & \mmcn{0.860/47.8} \\ \hline
				\mmc{(e)} & \mmc{Add BN} & \mmc{0.861/0.086} & \mmc{0.918/0.012} & \mmc{0.830/41.1} & \mmc{0.862/0.081} & \mmc{0.852/0.012} & \mmcn{0.824/50.7} \\ \hline				
			\end{tabular}
		}
	\end{center}
\end{table*}

\subsection{Domain Fidelity vs. Identity}
In Section \ref{sec:domainfidelity} and Section \ref{sec:ablation} of the main paper, we point out that merely using monotonicity loss can lead to a biased control, and present the domain fidelity loss $\mathcal{L}_{df}$. Here we provide more analysis on this setting and compare it with another naive alternative i.e., using a weak identity loss. As reported in Table \ref{table:MoreResults} (c), we first observe that adding an identity \emph{L1} loss on variant (b) does bring a wider control since it encourages the model to better preserve the fidelity of inputs, which otherwise is not modeled when merely using monotonicity constraint. However as a side effect of identity mapping, the GAN loss is thereby sacrificed and causes certain degradation on the control linearity. Further, through the comparison with our full model (a), using an identity loss also produces an inferior result, which on the other hand reveals the benefits of domain fidelity constraint, and supports our claims in the main paper.

\subsection{Impact of Different Normalization Methods}
We provide a theoretical analysis in Section \ref{sec:InsNorm} on the different normalization methods. As part of the model, we complete our paper with a more detailed analysis in Table \ref{table:MoreResults} (d-e). These models trained with additional normalization layers, generally, exhibit better smoothness yet with slightly inferior but acceptable translation quality compared with variant (a). Typically we find batch normalization (BN) leads to certain degradation on the linearity while IN on the contrary brings a boost. This is probably because IN better helps to model the style transition property in this task, while the inner contrastivity and dependency on training batch size narrow the improvements of BN. Overall, the choice of normalization layers serves as an option when implementing MonoPix, which brings certain differences but does not impede the monotonic modulation process.

\subsection{Visualization on the Activated Features}
\begin{figure*}[t!]
	\centering
		\begin{subfigure}[h]{\textwidth}
		\centering
		\resizebox{0.9\linewidth}{!}{
		\begin{tabular}{c c c c}

			\begin{overpic}[width=8em, valign=m]{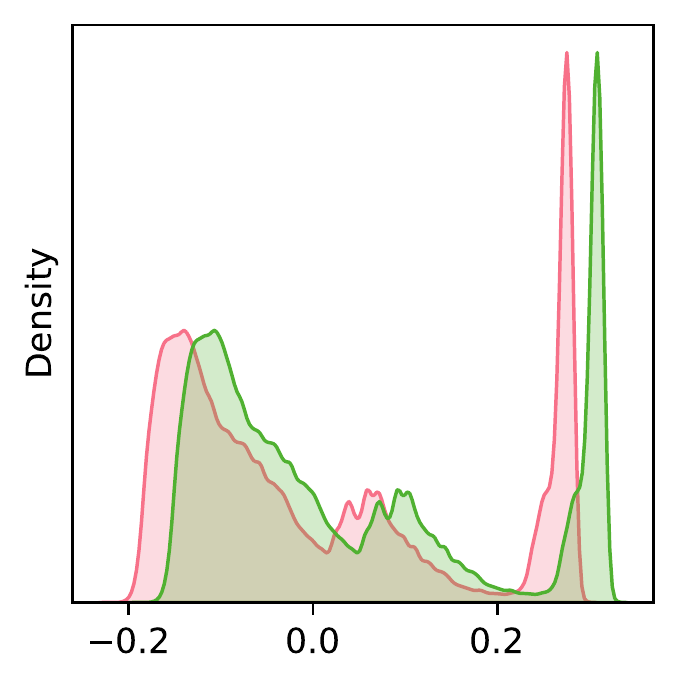}
				\put(15,90){\tiny\color{blue}{Input Distribution}}
			\end{overpic} &
			\begin{overpic}[width=8em, valign=m]{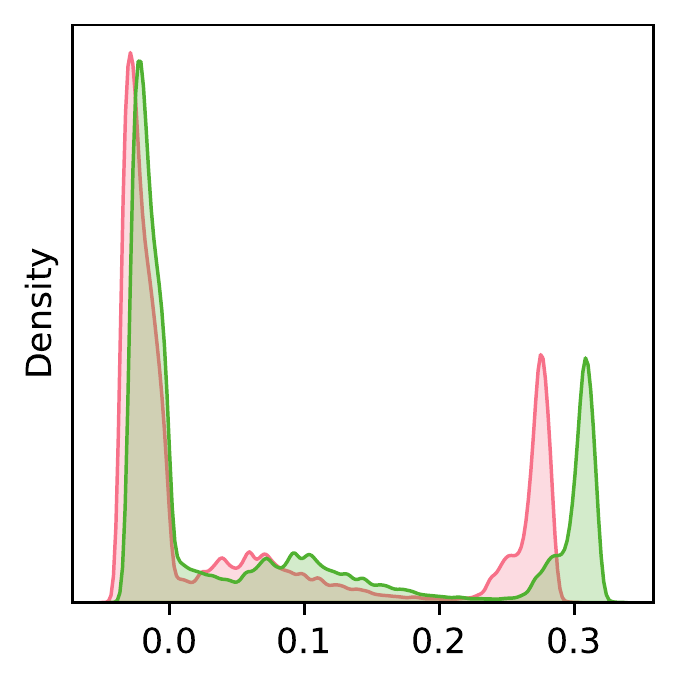}
				\put(15,90){\tiny\color{blue}{After Leaky ReLU}}
			\end{overpic}&
			\begin{overpic}[width=8em, valign=m]{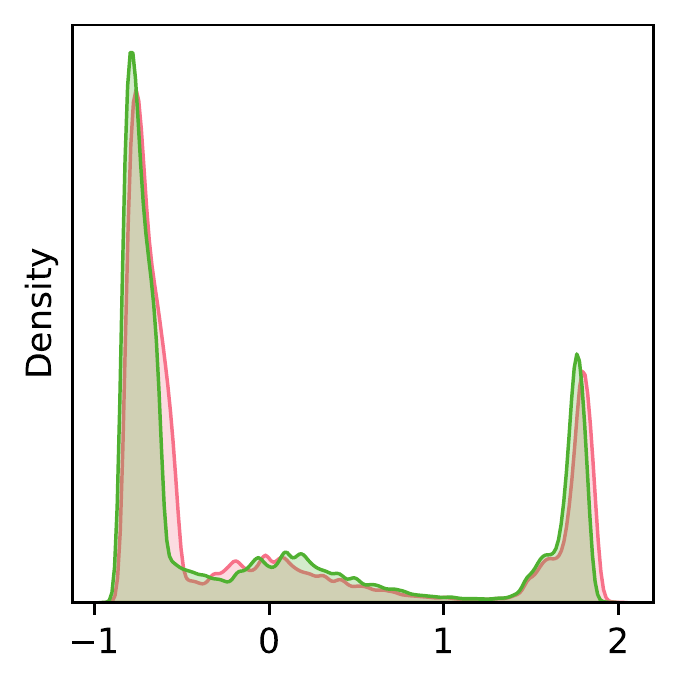}
				\put(30,90){\tiny\color{blue}{After IN}}
			\end{overpic}&
			\begin{overpic}[width=8em, valign=m]{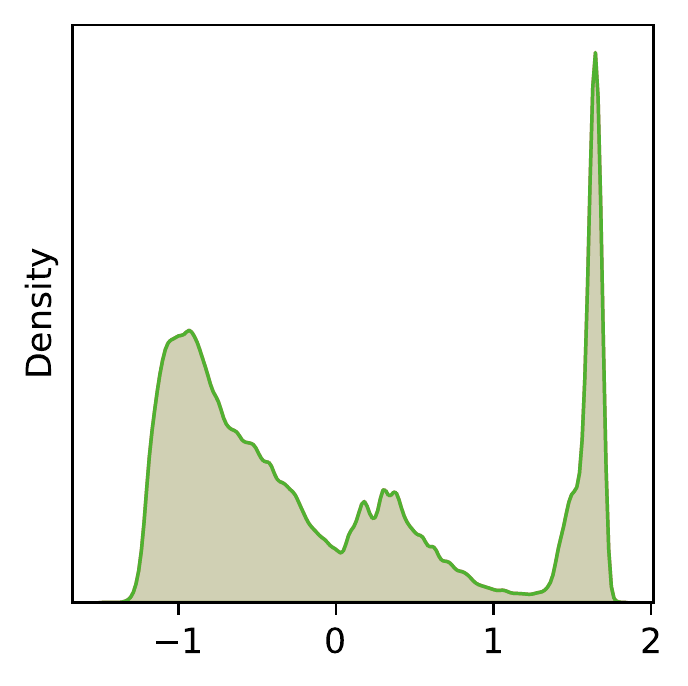}
				\put(30,90){\tiny\color{blue}{Plain IN}}
			\end{overpic}
		\end{tabular}}
		\caption{Adding a non-linear activation solves the IN dilemma. Red and green shadows denote the distribution of two feature maps obtained with varied modulation intensity}
		\end{subfigure}

		\begin{subfigure}[h]{\textwidth}
		\centering
		\resizebox{0.9\linewidth}{!}{
		\begin{tabular}{c c c c}
			\begin{overpic}[width=7em, valign=m]{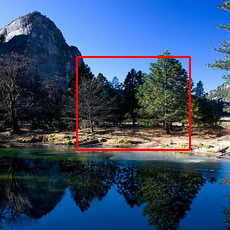}
				\put(35,90){\tiny\color{red}{Input}}
			\end{overpic}&
			\begin{overpic}[width=7em, valign=m]{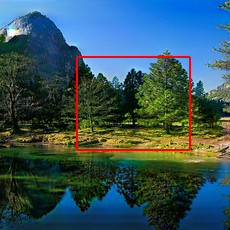}
				\put(25,90){\tiny\color{red}{Translated}}
			\end{overpic}&
			\begin{overpic}[width=7em, valign=m]{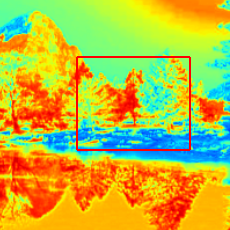}
				\put(10,90){\tiny\color{red}{Feature Intensity}}
			\end{overpic}&
			\begin{overpic}[width=7em, valign=m]{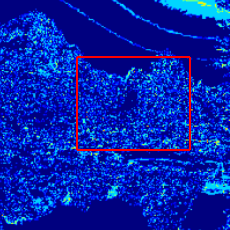}
				\put(10,90){\tiny\color{red}{Activated Regions}}
			\end{overpic}
		\end{tabular}}
		\caption{Visualization on the activated domain-specific features. The last but one picture evaluates the intensity of feature maps, while the rightmost picture provides visualization by locating where a non-linear transformation takes place}
		\end{subfigure}

	\caption{Illustration of the instance normalization-compatible modulation in MonoPix, and the corresponding visualizations}
	\label{fig:VisIN}

\end{figure*}

In the main paper, we propose to inject the contrastive controlling signal via a non-linear activation before adopting instance normalization (IN). Here we show in detail how this solution works in Figure \ref{fig:VisIN} (a), and the by-product of a visualization of activated regions in Figure \ref{fig:VisIN} (b). In the first sub-figure, we provide the numerical distribution of two feature maps obtained from different enhance levels of a same image. As can be found that a single non-linearity, combined with the convolution kernel, injects the intensity signal by ``pushing'' the feature distribution across non-linear activation regions and prevents a homogeneous representation after IN. Based on this property, we can further visualize the model's attention by examining where a non-linear transformation takes place (in leaky ReLU (LReLU) for example, we focus on the pixel entries that change their signs when the modulation intensity changes, as the non-linearity in LReLU is at zero), as we show in the rightmost picture in Figure \ref{fig:VisIN} (b). It provides a more precise visualization than simply relying on the absolute value of feature maps, whose result is provided in the last but one picture.

\subsection{Intensity of Interest}
\begin{figure*}[t!]
	\centering
	\resizebox{\linewidth}{!}{
		\setlength{\tabcolsep}{0.003\linewidth}
		\begin{tabular}{c c | c c c c c c c c c c c c}
			\includegraphics[width=4em, valign=m]{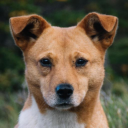}
			& ~
			& ~
			& \includegraphics[width=4em, valign=m]{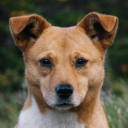}
			& \includegraphics[width=4em, valign=m]{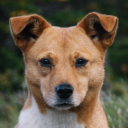}
			& \includegraphics[width=4em, valign=m]{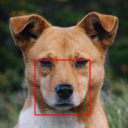}
			& \includegraphics[width=4em, valign=m]{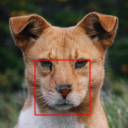}
			& \includegraphics[width=4em, valign=m]{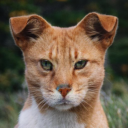}
			& \includegraphics[width=4em, valign=m]{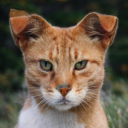}
			& \includegraphics[width=4em, valign=m]{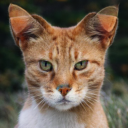}
			& \includegraphics[width=4em, valign=m]{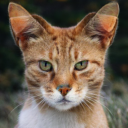}
			& \includegraphics[width=4em, valign=m]{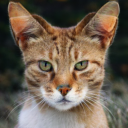}
			& \includegraphics[width=4em, valign=m]{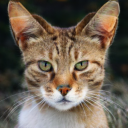}
			& \includegraphics[width=4em, valign=m]{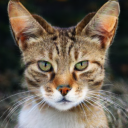}
			\\
			~&~&~&~&~&\tikzmark{lrs}{}&\tikzmark{rrs}{}&~&~&~&~&~&~&~ \\
			\tikzmark{a}{}&~&~&~&~&~&~&~&~&~&~&~&~&\tikzmark{b}{} 
			\\
			input & 
			~ &	 
			~ &	 
			$\boldsymbol{c}=0.0$ &
			$\boldsymbol{c}=0.1$ &
			$\boldsymbol{c}=0.2$ &
			$\boldsymbol{c}=0.3$ &
			$\boldsymbol{c}=0.4$ &
			$\boldsymbol{c}=0.5$ &
			$\boldsymbol{c}=0.6$ &
			$\boldsymbol{c}=0.7$ &
			$\boldsymbol{c}=0.8$ &
			$\boldsymbol{c}=0.9$ &
			$\boldsymbol{c}=1.0$ \\
			~&~&~&\tikzmark{lre}{}&~&~&~&~&~&~&~&~&~&\tikzmark{rre}{} \\
			N/A
			& ~ 
			& ~	
			& \includegraphics[width=4em, valign=m]{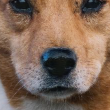}
			& \includegraphics[width=4em, valign=m]{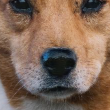}
			& \includegraphics[width=4em, valign=m]{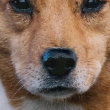}
			& \includegraphics[width=4em, valign=m]{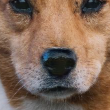}
			& \includegraphics[width=4em, valign=m]{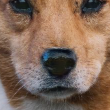}
			& \includegraphics[width=4em, valign=m]{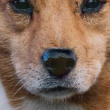}
			& \includegraphics[width=4em, valign=m]{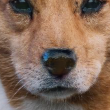}
			& \includegraphics[width=4em, valign=m]{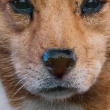}
			& \includegraphics[width=4em, valign=m]{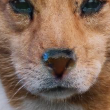}
			& \includegraphics[width=4em, valign=m]{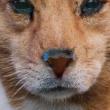}
			& \includegraphics[width=4em, valign=m]{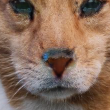}
			\\
			\tikzmark{c}{}&~&~&~&~&~&~&~&~&~&~&~&~&\tikzmark{d}{} 
			\\
			input & 
			~ &	 
			~ &	 
			$\boldsymbol{c}=0.20$ &
			$\boldsymbol{c}=0.21$ &
			$\boldsymbol{c}=0.22$ &
			$\boldsymbol{c}=0.23$ &
			$\boldsymbol{c}=0.24$ &
			$\boldsymbol{c}=0.25$ &
			$\boldsymbol{c}=0.26$ &
			$\boldsymbol{c}=0.27$ &
			$\boldsymbol{c}=0.28$ &
			$\boldsymbol{c}=0.29$ &
			$\boldsymbol{c}=0.30$ 
			\\
			
			\hline
			\\
			
			\includegraphics[width=4em, valign=m]{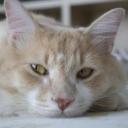}
			& ~
			& ~
			& \includegraphics[width=4em, valign=m]{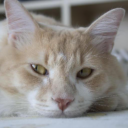}
			& \includegraphics[width=4em, valign=m]{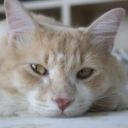}
			& \includegraphics[width=4em, valign=m]{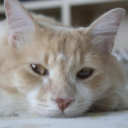}
			& \includegraphics[width=4em, valign=m]{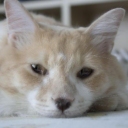}
			& \includegraphics[width=4em, valign=m]{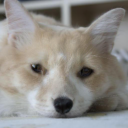}
			& \includegraphics[width=4em, valign=m]{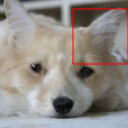}
			& \includegraphics[width=4em, valign=m]{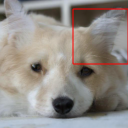}
			& \includegraphics[width=4em, valign=m]{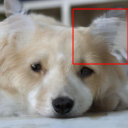}
			& \includegraphics[width=4em, valign=m]{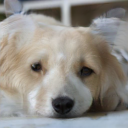}
			& \includegraphics[width=4em, valign=m]{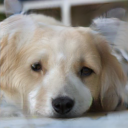}
			& \includegraphics[width=4em, valign=m]{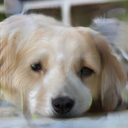}
			\\
			~&~&~&~&~&~&~&~&\tikzmark{lrs2}{}&~&\tikzmark{rrs2}{}&~&~&~ \\
			\tikzmark{a}{}&~&~&~&~&~&~&~&~&~&~&~&~&\tikzmark{b}{} 
			\\
			input & 
			~ &	 
			~ &	 
			$\boldsymbol{c}=0.0$ &
			$\boldsymbol{c}=0.1$ &
			$\boldsymbol{c}=0.2$ &
			$\boldsymbol{c}=0.3$ &
			$\boldsymbol{c}=0.4$ &
			$\boldsymbol{c}=0.5$ &
			$\boldsymbol{c}=0.6$ &
			$\boldsymbol{c}=0.7$ &
			$\boldsymbol{c}=0.8$ &
			$\boldsymbol{c}=0.9$ &
			$\boldsymbol{c}=1.0$ \\
			~&~&~&\tikzmark{lre2}{}&~&~&~&~&~&~&~&~&~&\tikzmark{rre2}{} \\
			N/A
			& ~ 
			& ~	
			& \includegraphics[width=4em, valign=m]{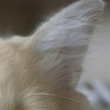}
			& \includegraphics[width=4em, valign=m]{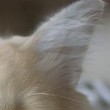}
			& \includegraphics[width=4em, valign=m]{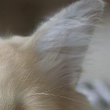}
			& \includegraphics[width=4em, valign=m]{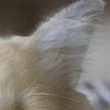}
			& \includegraphics[width=4em, valign=m]{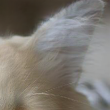}
			& \includegraphics[width=4em, valign=m]{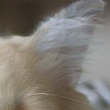}
			& \includegraphics[width=4em, valign=m]{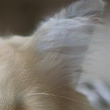}
			& \includegraphics[width=4em, valign=m]{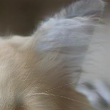}
			& \includegraphics[width=4em, valign=m]{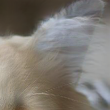}
			& \includegraphics[width=4em, valign=m]{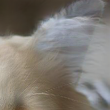}
			& \includegraphics[width=4em, valign=m]{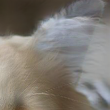}
			\\
			\tikzmark{c}{}&~&~&~&~&~&~&~&~&~&~&~&~&\tikzmark{d}{} 
			\\
			input & 
			~ &	 
			~ &	 
			$\boldsymbol{c}=0.50$ &
			$\boldsymbol{c}=0.52$ &
			$\boldsymbol{c}=0.54$ &
			$\boldsymbol{c}=0.56$ &
			$\boldsymbol{c}=0.58$ &
			$\boldsymbol{c}=0.60$ &
			$\boldsymbol{c}=0.62$ &
			$\boldsymbol{c}=0.64$ &
			$\boldsymbol{c}=0.66$ &
			$\boldsymbol{c}=0.68$ &
			$\boldsymbol{c}=0.70$
 

		\end{tabular}\linkdash{lrs}{lre}\linkdash{rrs}{rre}\linkdash{lrs2}{lre2}\linkdash{rrs2}{rre2}} 
	\caption{Illustration of Intensity of interest (IOI), where $\boldsymbol{c}$ denotes the translation intensity. A fine-grained continuous translation can be obtained on intensities of interest selectively, for example to observe how whiskers and cats' eyes grow (the second row) and how a cat's ear evolves to that of a dog (the last row)}
	\label{fig:IOI}\end{figure*}

As we mentioned in Section \ref{sec:res_domaintrans} of the main paper, \name exhibits a considerably better absolute linearity which shows the translation is loyal and monotonic, but the adjacent modifications are not perfectly and evenly distributed. 
We note that this is not a practical problem since we can, actually, provide a finer and monotonic modulation among ``intensity of interest'' instead of relying on uniformly sampled intensities. In real scenarios, it is also expected in most cases that a user modulates in a selective and attentive manner (such as employing ternary search), 
where a high absolute linearity may count more than relative linearity (a low absolute linearity incurs risk of non-unimodal aesthetic curve, while a low relative linearity will not impede ternary search but degrades the complexity). 
In Figure \ref{fig:IOI}, we take the growth of whiskers and cats' eyes, and evolution of dog ears as examples to better demonstrate this point. A finer and detailed modulation is provided next to the overall modulations.

\def\arraystretch{0.5}

\subsection{Generalization on Latent Representation}
\begin{figure*}[t!]
	\centering
	\resizebox{\linewidth}{!}{
		\setlength{\tabcolsep}{0.003\linewidth}
		
		\begin{tabular}{c c | c c c c c c c}
			\includegraphics[width=6em, valign=m]{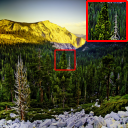}
			& ~
			& ~
			& \includegraphics[width=6em, valign=m]{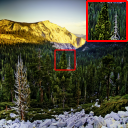}
			& \includegraphics[width=6em, valign=m]{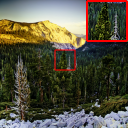}
			& \includegraphics[width=6em, valign=m]{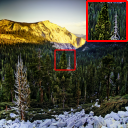}
			& \includegraphics[width=6em, valign=m]{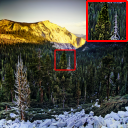}
			& \includegraphics[width=6em, valign=m]{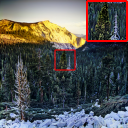}
			& \includegraphics[width=6em, valign=m]{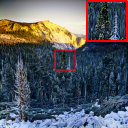}
			\\
			\tikzmark{a}{}&~&~&~&~&~&~&~&~\tikzmark{b}{} \\
			input & 
			~ &	 
			~ &	 
			$\boldsymbol{c}=0.0$ &
			$\boldsymbol{c}=0.2$ &
			$\boldsymbol{c}=0.4$ &
			$\boldsymbol{c}=0.6$ &
			$\boldsymbol{c}=0.8$ &
			$\boldsymbol{c}=1.0$ \\

			\includegraphics[width=6em, valign=m]{FigSupp/OOB/89_enhlvl_99.png}
			& ~ 
			& ~	
			& \includegraphics[width=6em, valign=m]{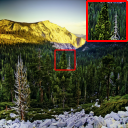}
			& \includegraphics[width=6em, valign=m]{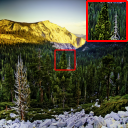}
			& \includegraphics[width=6em, valign=m]{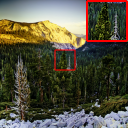}
			& \includegraphics[width=6em, valign=m]{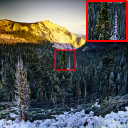}
			& \includegraphics[width=6em, valign=m]{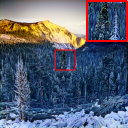}
			& \includegraphics[width=6em, valign=m]{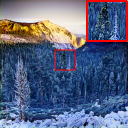}
			\\
			\tikzmark{c}{}&~&~&~&~&~&~&~&~\tikzmark{d}{} 
			\\
			input & 
			~ &	 
			~ &	 
			$\boldsymbol{c}=-1.0$ &
			$\boldsymbol{c}=-0.4$ &
			$\boldsymbol{c}=0.2$ &
			$\boldsymbol{c}=0.8$ &
			$\boldsymbol{c}=1.4$ &
			$\boldsymbol{c}=2.0$ 
			\\

		\end{tabular}\link{c}{d}\link{a}{b}}
	\caption{Out-of-the-bound (OOB) generalization. MonoPix retains its robustness and generates realistic domain-specific images in a wider control range}
	\label{fig:OOB}

\end{figure*}

Besides the main results, it is critical to examine the generalization ability of a learned generator. In previous works, it is evidenced that a well-learned generator retains the robustness and generates meaningful results by style interpolation \cite{lee2018diverse}, feeding unseen conditional inputs \cite{reed2016generative}, or out-of-the-bound inference \cite{chen2016infogan}. In the main paper, we have provided a detailed explanation on how a global intensity training enables pixel-level manipulation in the inference stage, which can be viewed as a spatial-level generalization. Here we provide more examples on the out-of-the-bound (OOB) manipulation, which is only mentioned in Section \ref{sec:oob} (in the main paper) but not specifically presented. As can be found in Figure \ref{fig:OOB}, MonoPix is compatible with a slightly OOB inference and generates natural-looking and acceptable results. 
Specifically, modulation with a negative modulation intensity can usually yield a better fidelity while those with higher intensities more resemble the target domain. This evidence suggests that the derived model does not merely memorize the input intensities, but learns to represent them in a linear and smooth latent space.  

\def\arraystretch{0.5}

\section{More Qualitative Evaluations}

\subsection{Image Quality over the Modulation Trajectory}
\label{sec:qualityover}
\begin{figure*}[b!]
	\centering
	\resizebox{\linewidth}{!}{
		\setlength{\tabcolsep}{0.003\linewidth}
		
		\begin{tabular}{c c | c c c c c c c c c c c c}
			\includegraphics[width=6em, valign=m]{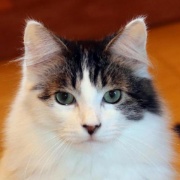}
			& ~
			& ~
			& \includegraphics[width=6em, valign=m]{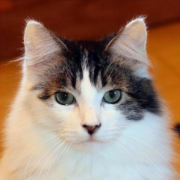} 
			& \includegraphics[width=6em, valign=m]{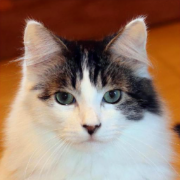} 
			& \includegraphics[width=6em, valign=m]{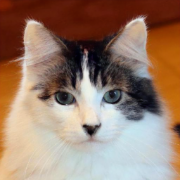} 
			& \includegraphics[width=6em, valign=m]{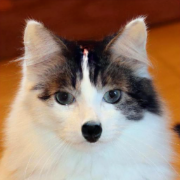} 
			& \includegraphics[width=6em, valign=m]{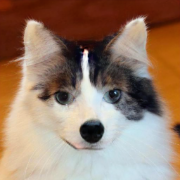} 
			& \includegraphics[width=6em, valign=m]{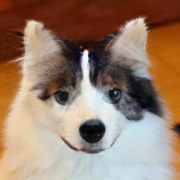} 
			& \includegraphics[width=6em, valign=m]{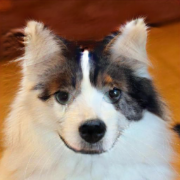} 
			& \includegraphics[width=6em, valign=m]{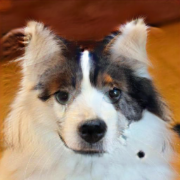} 
			& \includegraphics[width=6em, valign=m]{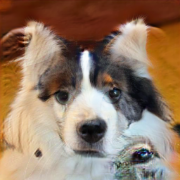} 
			& \includegraphics[width=6em, valign=m]{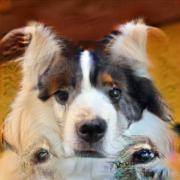} 
			& \includegraphics[width=6em, valign=m]{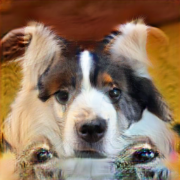} \\
			\tikzmark{a}{}&~&~&~&~&~&~&~&~&~&~&~&~&\tikzmark{b}{} \\
			input & 
			~ &	 
			~ &	 
			$\boldsymbol{c}=0.0$ &
			$\boldsymbol{c}=0.1$ &
			$\boldsymbol{c}=0.2$ &
			$\boldsymbol{c}=0.3$ &
			$\boldsymbol{c}=0.4$ &
			$\boldsymbol{c}=0.5$ &
			$\boldsymbol{c}=0.6$ &
			$\boldsymbol{c}=0.7$ &
			$\boldsymbol{c}=0.8$ &
			$\boldsymbol{c}=0.9$ &
			$\boldsymbol{c}=1.0$ \\
		
		\end{tabular}\link{a}{b}}
	\caption{Discrepancy between single-image and overall quality. \name does not assume that images with the highest translation intensities are best candidates}
	\label{fig:Unnatural}

\end{figure*}

One major difference between \name and prevalent generative models is that, \name does not assume $\boldsymbol{c}=1$ is the optimal translation intensity, instead it learns through relative ``reward'' from domain discriminator. As a result, the optimal translation intensities vary for different input samples and are determined (hopefully) by ternary search. An example can be found in Figure \ref{fig:Unnatural}. Though we are able to collect the highest accuracy (ACC) across the whole generation trajectory, calculating FID on a specific, and fixed intensity for \name can lead to an underestimation. In Table \ref{table:morefid}, we provide an example of evaluating FID on the last translation intensity $\boldsymbol{c}=1$, from which we can observe a relatively poor performance than on the whole trajectory. Nevertheless, we notice that evaluating FID on the whole, evenly-spaced intensities may not provide an ideal metric, as it discourages the model from exploring mixed-style intermediate samples, which can be identified from a poor relative linearity.

\def\arraystretch{0.5}

\def\arraystretch{1}
\begin{table}[h]
	\begin{center}
		\caption{More FID comparisons. We report FID with the highest translation intensity (last modulation result) and across the whole trajectory (overall). Translation from Summer to Winter is denoted as S2W, and Cat to Dog as C2D}
		\label{table:morefid}
		\scalebox{1.0}{
			\begin{tabular}{lllllllll}
				\hline
				\mmc{~} & \multicolumn{4}{c|}{last} & \multicolumn{4}{c}{overall} \\ \cline{1-9}
				\mmc{Directions} &\mmc{S2W}&\mmc{W2S}&\mmc{C2D}&\mmc{D2C}&\mmc{S2W}&\mmc{W2S}&\mmc{C2D}&\mmcn{D2C} \\ \hline
				\mmc{CycleGAN-DNI \cite{zhu2017unpaired}}&\mmc{60.2}&\mmc{66.0}&\mmc{-}&\mmc{-}&\mmc{35.0}&\mmc{42.8}&\mmc{-}&\mmcn{-} \\ \hline
				\mmc{StarGANv2-Rdm \cite{choi2020stargan}}&\mmc{71.7}&\mmc{68.9}&\mmc{53.9}&\mmc{17.7}&\mmc{46.1}&\mmc{50.1}&\mmc{33.9}&\mmcn{25.7} \\ \hline
				\mmc{StarGANv2-Cent \cite{choi2020stargan}}&\mmc{78.8}&\mmc{67.7}&\mmc{115}&\mmc{69.4}&\mmc{49.3}&\mmc{52.8}&\mmc{50.2}&\mmcn{38.0} \\ \hline
				\mmc{Liu \etal -Rdm \cite{liu2021smoothing}}&\mmc{-}&\mmc{-}&\mmc{72.8}&\mmc{23.1}&\mmc{-}&\mmc{-}&\mmc{37.1}&\mmcn{27.5}\\ \hline
				\mmc{SAVI2I-Rdm \cite{mao2022continuous}}&\mmc{70.7}&\mmc{74.1}&\mmc{44.1}&\mmc{16.6}&\mmc{44.5}&\mmc{47.6}&\mmc{28.6}&\mmcn{24.2} \\ \hline
				\mmc{SAVI2I-Cent \cite{mao2022continuous}}&\mmc{75.9}&\mmc{70.1}&\mmc{158}&\mmc{63.7}&\mmc{47.5}&\mmc{49.2}&\mmc{53.6}&\mmcn{31.4} \\ \hline
				\mmc{MonoPix}&\mmc{63.5}&\mmc{74.7}&\mmc{87.4}&\mmc{27.1}&\mmc{38.1}&\mmc{47.3}&\mmc{41.1}&\mmcn{14.4}\\ \hline
			\end{tabular}
		}
	\end{center}
\end{table}

\subsection{Subjective Consistency Analysis}
We further carry out subjective consistency analysis to both check the tenability of our metrics and provide complementary evaluations. The results are listed in Table \ref{table:Subjective}. We first notice that the \textit{linearity}, \textit{fidelity}, and \textit{smoothness} metrics are fairly consistent with quantitative scores in the main paper. However, the quality of translation (\ie, \textit{success} and \textit{quality}) differs in Yosemite and AFHQ datasets. The primary reason behind this lies in the fact that \name is now integrated with a CycleGAN-style translation framework, where the shape-related attributes are not particularly taken into consideration (compared with StarGANv2). Consequently, \name shows a better fidelity and low-level style manipulation capability, but may not behave well in multi-attribute modulations. We also notice that the performance degradation is also related to the property we mention in \ref{sec:qualityover} (in the supplementary material. Also in Section \ref{sec:res_domaintrans} of the main paper), where in some cases \name provides unrealistic images near $\boldsymbol{c}=1$. Though evaluating FID (typically on the whole trajectory) does not particularly reveal this issue, human critics are more sensitive to the final translated results.

\def\arraystretch{1}
\begin{table}[h]
	\begin{center}
		\caption{User study on \name versus SAVI2I and StarGANv2. We show user preferences in percentages}
		\label{table:Subjective}
		\scalebox{1.0}{
			\begin{tabular}{ccccc}
				\hline
				\mmc{~} & \multicolumn{2}{c|}{Yosemite} & \multicolumn{2}{c}{AFHQ} \\ \hline
				\mmc{Queries} &\mmcc{vs. SAVI2I}&\mmc{vs. StarGANv2}&\mmc{vs. SAVI2I}&\mmcn{vs. StarGANv2} \\ \hline
				\mmc{Linearity}&\mmcc{43.3}&\mmcc{48.0}&\mmcc{41.3}&\mmccn{70.2} \\ \hline
				\mmc{Fidelity}&\mmcc{84.0}&\mmcc{84.7}&\mmcc{81.9}&\mmccn{81.0} \\ \hline
				\mmc{Smoothness}&\mmcc{60.0}&\mmcc{56.8}&\mmcc{78.4}&\mmccn{87.9} \\ \hline
				\mmc{Success}&\mmcc{63.2}&\mmcc{60.4}&\mmcc{27.4}&\mmccn{30.4} \\ \hline
				\mmc{Quality}&\mmcc{65.9}&\mmcc{67.4}&\mmcc{26.3}&\mmccn{37.9} \\ \hline
			\end{tabular}
		}
	\end{center}
\end{table}

\clearpage
\section{More Visual Results}

\begin{figure*}[h!]
	\centering
	\resizebox{\linewidth}{!}{
		\setlength{\tabcolsep}{0.003\linewidth}
		\begin{tabular}{c c c c | c c c c c c c c c c c c c}
			\multicolumn{15}{c}{Winter $\mapsto$ Summer}\\
			\multirow{1}{*}[0.8cm]{\rotatebox{90}{\tiny{CycleGAN-DNI}}} & \multirow{1}{*}[0.35cm]{\rotatebox{90}{\tiny \cite{zhu2017unpaired,wang2019deep}}} 
			& \includegraphics[width=5em, valign=m]{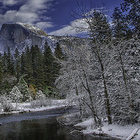}
			& ~
			& ~
			& \includegraphics[width=5em, valign=m]{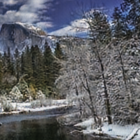}
			& \includegraphics[width=5em, valign=m]{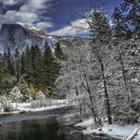}
			& \includegraphics[width=5em, valign=m]{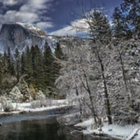}
			& \includegraphics[width=5em, valign=m]{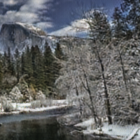}
			& \includegraphics[width=5em, valign=m]{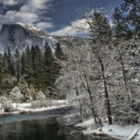}
			& \includegraphics[width=5em, valign=m]{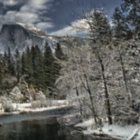}
			& \includegraphics[width=5em, valign=m]{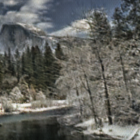}
			& \includegraphics[width=5em, valign=m]{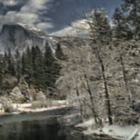}
			& \includegraphics[width=5em, valign=m]{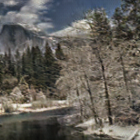}
			& \includegraphics[width=5em, valign=m]{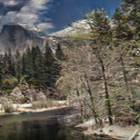}
			& \includegraphics[width=5em, valign=m]{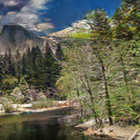}
			\\
			\multirow{1}{*}[0.55cm]{\rotatebox{90}{\tiny{StarGANv2}}} & \rotatebox{90}{\tiny \cite{choi2020stargan}} 
			& \includegraphics[width=5em, valign=m]{FigSupp/MoreVis/W2S/CycleganDNI/161_enhlvl_99.png}
			& ~ 
			& ~	
			& \includegraphics[width=5em, valign=m]{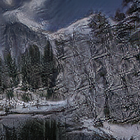}
			& \includegraphics[width=5em, valign=m]{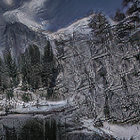}
			& \includegraphics[width=5em, valign=m]{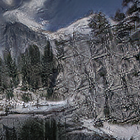}
			& \includegraphics[width=5em, valign=m]{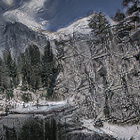}
			& \includegraphics[width=5em, valign=m]{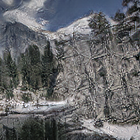}
			& \includegraphics[width=5em, valign=m]{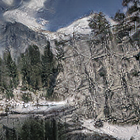}
			& \includegraphics[width=5em, valign=m]{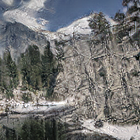}
			& \includegraphics[width=5em, valign=m]{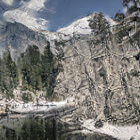}
			& \includegraphics[width=5em, valign=m]{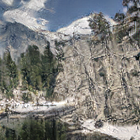}
			& \includegraphics[width=5em, valign=m]{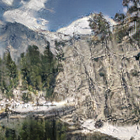}
			& \includegraphics[width=5em, valign=m]{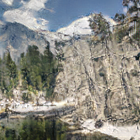}
			\\
			\multirow{1}{*}[0.35cm]{\rotatebox{90}{\tiny{SAVI2I}}} & \rotatebox{90}{\tiny \cite{mao2022continuous} }
			& \includegraphics[width=5em, valign=m]{FigSupp/MoreVis/W2S/CycleganDNI/161_enhlvl_99.png}
			& ~ 
			& ~	
			& \includegraphics[width=5em, valign=m]{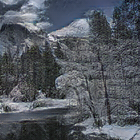}
			& \includegraphics[width=5em, valign=m]{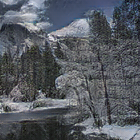}
			& \includegraphics[width=5em, valign=m]{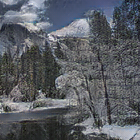}
			& \includegraphics[width=5em, valign=m]{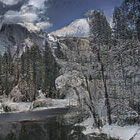}
			& \includegraphics[width=5em, valign=m]{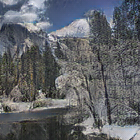}
			& \includegraphics[width=5em, valign=m]{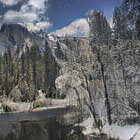}
			& \includegraphics[width=5em, valign=m]{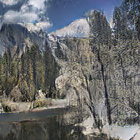}
			& \includegraphics[width=5em, valign=m]{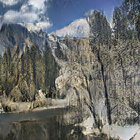}
			& \includegraphics[width=5em, valign=m]{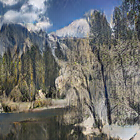}
			& \includegraphics[width=5em, valign=m]{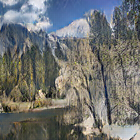}
			& \includegraphics[width=5em, valign=m]{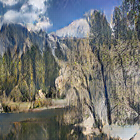}
			\\
			\multirow{1}{*}[0.45cm]{\rotatebox{90}{\tiny{MonoPix}}} & \multirow{1}{*}[0.3cm]{\rotatebox{90}{\tiny{(Ours)}}}
			& \includegraphics[width=5em, valign=m]{FigSupp/MoreVis/W2S/CycleganDNI/161_enhlvl_99.png}
			& ~ 
			& ~	
			& \includegraphics[width=5em, valign=m]{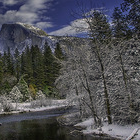}
			& \includegraphics[width=5em, valign=m]{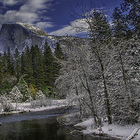}
			& \includegraphics[width=5em, valign=m]{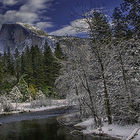}
			& \includegraphics[width=5em, valign=m]{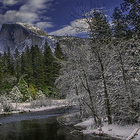}
			& \includegraphics[width=5em, valign=m]{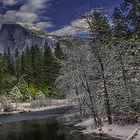}
			& \includegraphics[width=5em, valign=m]{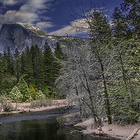}
			& \includegraphics[width=5em, valign=m]{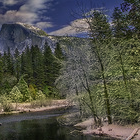}
			& \includegraphics[width=5em, valign=m]{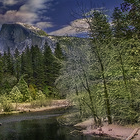}
			& \includegraphics[width=5em, valign=m]{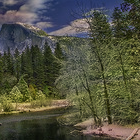}
			& \includegraphics[width=5em, valign=m]{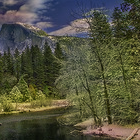}
			& \includegraphics[width=5em, valign=m]{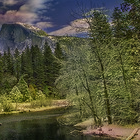}
			\\
			\multicolumn{15}{c}{Dog $\mapsto$ Cat}\\
			\multirow{1}{*}[0.5cm]{\rotatebox{90}{\tiny{StarGANv2}}} & \rotatebox{90}{\tiny \cite{choi2020stargan}}
			& \includegraphics[width=5em, valign=m]{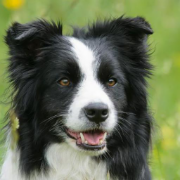}
			& ~ 
			& ~	
			& \includegraphics[width=5em, valign=m]{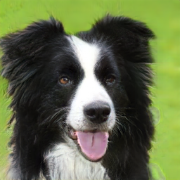}
			& \includegraphics[width=5em, valign=m]{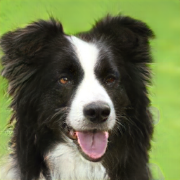}
			& \includegraphics[width=5em, valign=m]{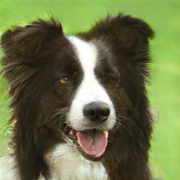}
			& \includegraphics[width=5em, valign=m]{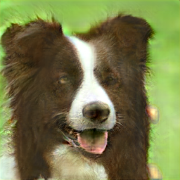}
			& \includegraphics[width=5em, valign=m]{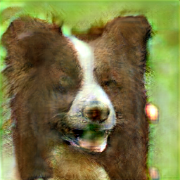}
			& \includegraphics[width=5em, valign=m]{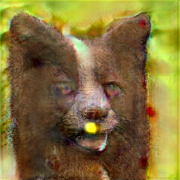}
			& \includegraphics[width=5em, valign=m]{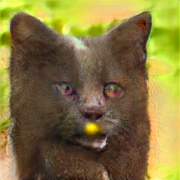}
			& \includegraphics[width=5em, valign=m]{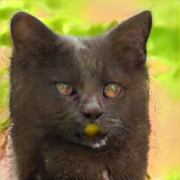}
			& \includegraphics[width=5em, valign=m]{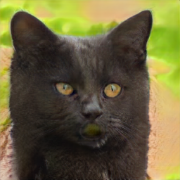}
			& \includegraphics[width=5em, valign=m]{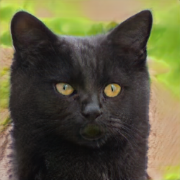}
			& \includegraphics[width=5em, valign=m]{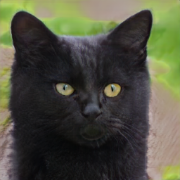}
			\\
			\multirow{1}{*}[0.4cm]{\rotatebox{90}{\tiny{Liu \etal}}} & \rotatebox{90}{\tiny \cite{liu2021smoothing}}
			& \includegraphics[width=5em, valign=m]{FigSupp/MoreVis/D2C/MonoPix/348_enhlvl_99.png}
			& ~ 
			& ~	
			& \includegraphics[width=5em, valign=m]{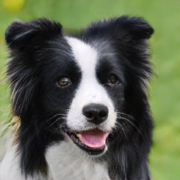}
			& \includegraphics[width=5em, valign=m]{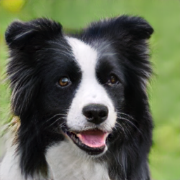}
			& \includegraphics[width=5em, valign=m]{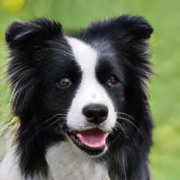}
			& \includegraphics[width=5em, valign=m]{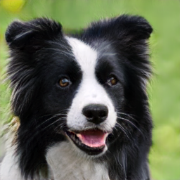}
			& \includegraphics[width=5em, valign=m]{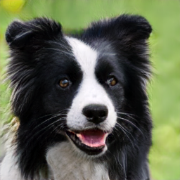}
			& \includegraphics[width=5em, valign=m]{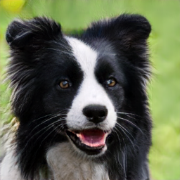}
			& \includegraphics[width=5em, valign=m]{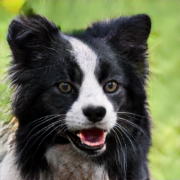}
			& \includegraphics[width=5em, valign=m]{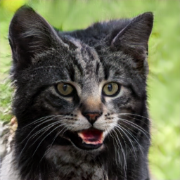}
			& \includegraphics[width=5em, valign=m]{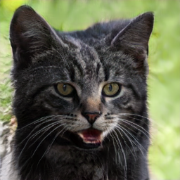}
			& \includegraphics[width=5em, valign=m]{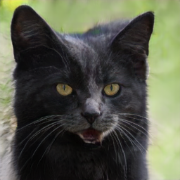}
			& \includegraphics[width=5em, valign=m]{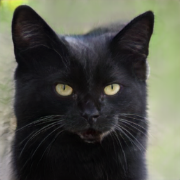}
			\\
			\multirow{1}{*}[0.3cm]{\rotatebox{90}{\tiny{SAVI2I}}} & \rotatebox{90}{\tiny \cite{mao2022continuous}}
			& \includegraphics[width=5em, valign=m]{FigSupp/MoreVis/D2C/MonoPix/348_enhlvl_99.png}
			& ~ 
			& ~	
			& \includegraphics[width=5em, valign=m]{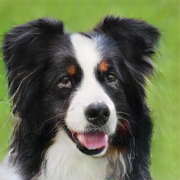}
			& \includegraphics[width=5em, valign=m]{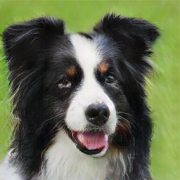}
			& \includegraphics[width=5em, valign=m]{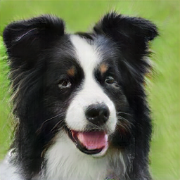}
			& \includegraphics[width=5em, valign=m]{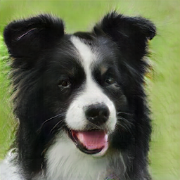}
			& \includegraphics[width=5em, valign=m]{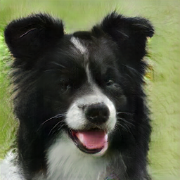}
			& \includegraphics[width=5em, valign=m]{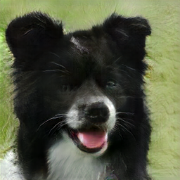}
			& \includegraphics[width=5em, valign=m]{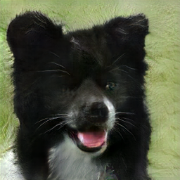}
			& \includegraphics[width=5em, valign=m]{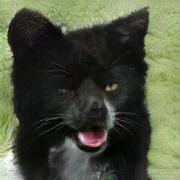}
			& \includegraphics[width=5em, valign=m]{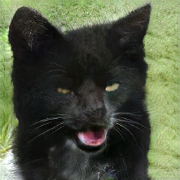}
			& \includegraphics[width=5em, valign=m]{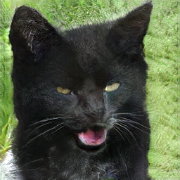}
			& \includegraphics[width=5em, valign=m]{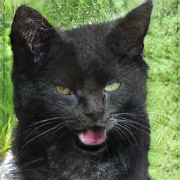}
			\\
			
			\multirow{1}{*}[0.35cm]{\rotatebox{90}{\tiny{MonoPix}}} & \multirow{1}{*}[0.3cm]{\rotatebox{90}{\tiny{(Ours)}}}
			& \includegraphics[width=5em, valign=m]{FigSupp/MoreVis/D2C/MonoPix/348_enhlvl_99.png}
			& ~ 
			& ~	
			& \includegraphics[width=5em, valign=m]{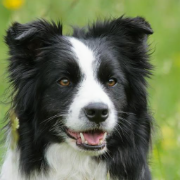}
			& \includegraphics[width=5em, valign=m]{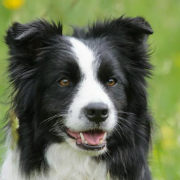}
			& \includegraphics[width=5em, valign=m]{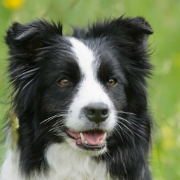}
			& \includegraphics[width=5em, valign=m]{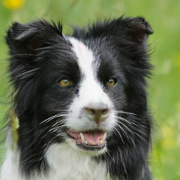}
			& \includegraphics[width=5em, valign=m]{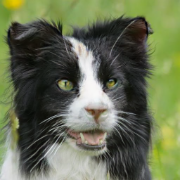}
			& \includegraphics[width=5em, valign=m]{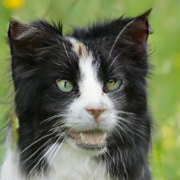}
			& \includegraphics[width=5em, valign=m]{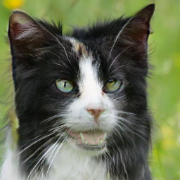}
			& \includegraphics[width=5em, valign=m]{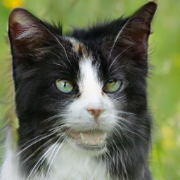}
			& \includegraphics[width=5em, valign=m]{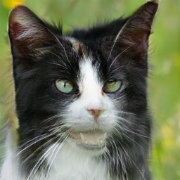}
			& \includegraphics[width=5em, valign=m]{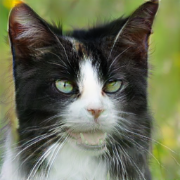}
			& \includegraphics[width=5em, valign=m]{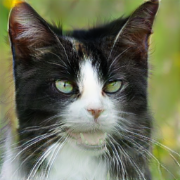}
			\\
			~ & ~ & \tikzmark{c}{}&~&~&~&~&~&~&~&~&~&~&~&~&\tikzmark{d}{} 
			\\
			~ & ~ &
			input & 
			~ &	 
			~ &	 
			$\boldsymbol{c}=0.0$ &
			$\boldsymbol{c}=0.1$ &
			$\boldsymbol{c}=0.2$ &
			$\boldsymbol{c}=0.3$ &
			$\boldsymbol{c}=0.4$ &
			$\boldsymbol{c}=0.5$ &
			$\boldsymbol{c}=0.6$ &
			$\boldsymbol{c}=0.7$ &
			$\boldsymbol{c}=0.8$ &
			$\boldsymbol{c}=0.9$ &
			$\boldsymbol{c}=1.0$ &
		\end{tabular}\link{c}{d}}
	\caption{More qualitative results on Yosemite winter to summer and AFHQ dog to cat translation}
	\label{fig:MoreDomaonTranslation}
\end{figure*}

\begin{figure*}[t!]
	\centering
	\resizebox{\linewidth}{!}{
		\setlength{\tabcolsep}{0.003\linewidth}
		\large
		\begin{tabular}{c c c c c c}
			\multicolumn{6}{c}{\small Low light $\mapsto$ Normal light}\\
			\includegraphics[width=6em, valign=m]{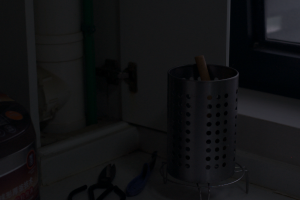}
			& \includegraphics[width=6em, valign=m]{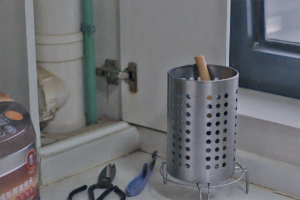}
			& \includegraphics[width=6em, valign=m]{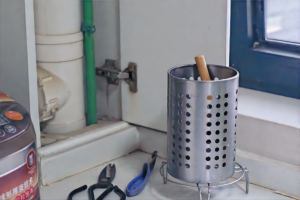}
			& \includegraphics[width=6em, valign=m]{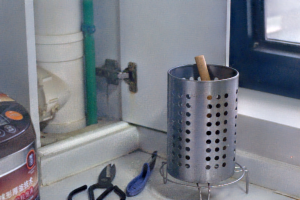}
			& \includegraphics[width=6em, valign=m]{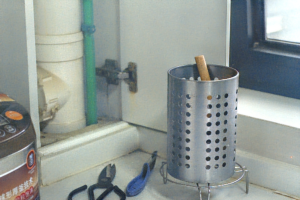}
			& \includegraphics[width=6em, valign=m]{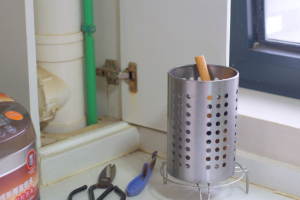}
			\\
			\includegraphics[width=6em, valign=m]{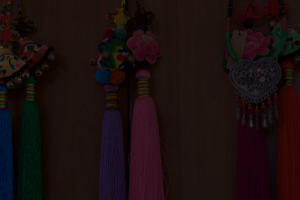}
			& \includegraphics[width=6em, valign=m]{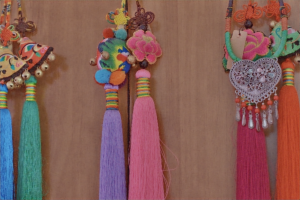}
			& \includegraphics[width=6em, valign=m]{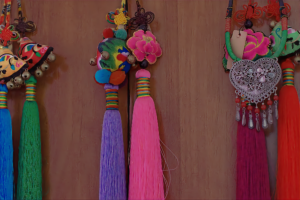}
			& \includegraphics[width=6em, valign=m]{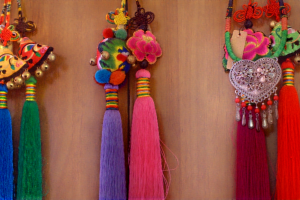}
			& \includegraphics[width=6em, valign=m]{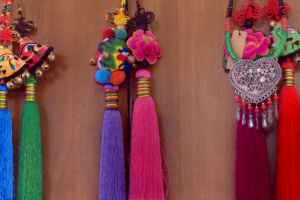}
			& \includegraphics[width=6em, valign=m]{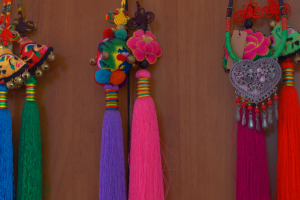}
			\\
			\includegraphics[width=6em, valign=m]{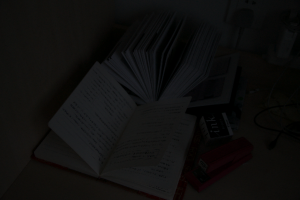}
			& \includegraphics[width=6em, valign=m]{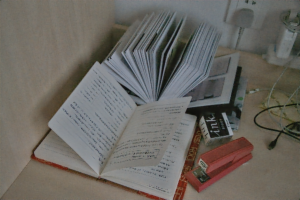}
			& \includegraphics[width=6em, valign=m]{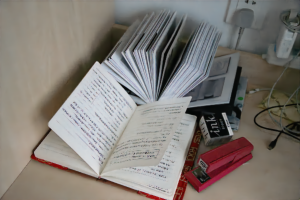}
			& \includegraphics[width=6em, valign=m]{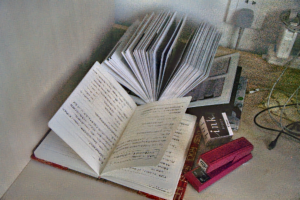}
			& \includegraphics[width=6em, valign=m]{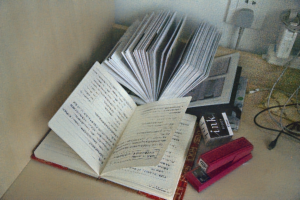}
			& \includegraphics[width=6em, valign=m]{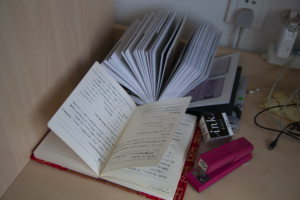}
			\\
			\\ ~
			\small Input &
			\small TBEFN \cite{lu2020tbefn} &
			\small KinD++ Ref \cite{zhang2021beyond} &
			\small EnlightenGAN \cite{jiang2021enlightengan} &
			\small MonoPix TS &
			\small Reference

			\\ ~ 
			\\\multicolumn{6}{c}{\small Clean $\mapsto$ Noisy}\\
			\includegraphics[width=5em, valign=m]{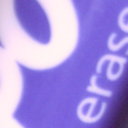}
			& \includegraphics[width=5em, valign=m]{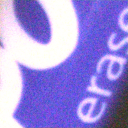}
			& \includegraphics[width=5em, valign=m]{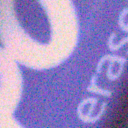}
			& \includegraphics[width=5em, valign=m]{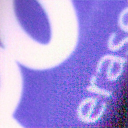}
			& \includegraphics[width=5em, valign=m]{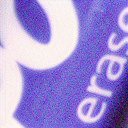}
			& \includegraphics[width=5em, valign=m]{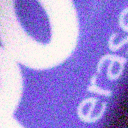}
			\\
			\includegraphics[width=5em, valign=m]{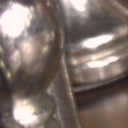}
			& \includegraphics[width=5em, valign=m]{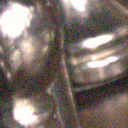}
			& \includegraphics[width=5em, valign=m]{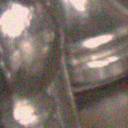}
			& \includegraphics[width=5em, valign=m]{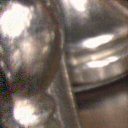}
			& \includegraphics[width=5em, valign=m]{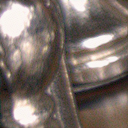}
			& \includegraphics[width=5em, valign=m]{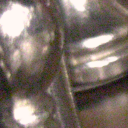}
			\\
			\includegraphics[width=5em, valign=m]{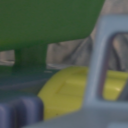}
			& \includegraphics[width=5em, valign=m]{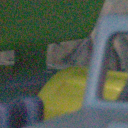}
			& \includegraphics[width=5em, valign=m]{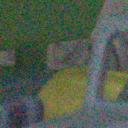}
			& \includegraphics[width=5em, valign=m]{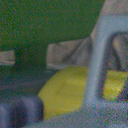}
			& \includegraphics[width=5em, valign=m]{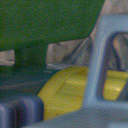}
			& \includegraphics[width=5em, valign=m]{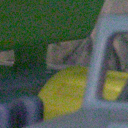}
			\\ ~
			\\
			\small Input &
			\small CBDNet \cite{guo2019toward} &
			\small ULRD \cite{brooks2019unprocessing} &
			\small DANet \cite{yue2020dual} &
			\small MonoPix TS &
			\small Reference
			
		\end{tabular}}
	\caption{Visual comparisons on the LOL low-light enhancement task and SIDD natural noise generation. In MonoPix, ``TS'' denotes ``ternary search''}
	\label{fig:MoreLowLevel}
\end{figure*}

\begin{figure*}[t!]
	\centering
	\resizebox{\linewidth}{!}{
		\setlength{\tabcolsep}{0.003\linewidth}
		\large
		\begin{tabular}{c c | c c c c c}
			\includegraphics[width=6em, valign=m]{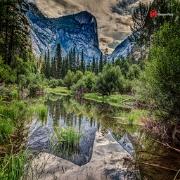}
			& ~
			& ~
			& \includegraphics[width=6em, valign=m]{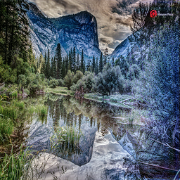}
			& \includegraphics[width=6em, valign=m]{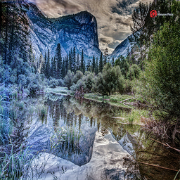}
			& \includegraphics[width=6em, valign=m]{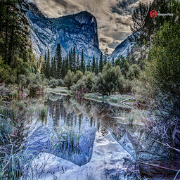}
			& \includegraphics[width=6em, valign=m]{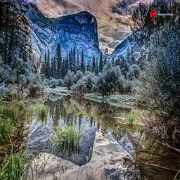}
			\\
			\includegraphics[width=6em, valign=m]{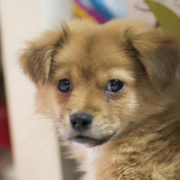}
			& ~
			& ~
			& \includegraphics[width=6em, valign=m]{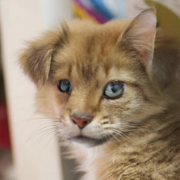}
			& \includegraphics[width=6em, valign=m]{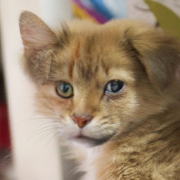}
			& \includegraphics[width=6em, valign=m]{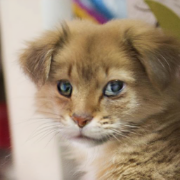}
			& \includegraphics[width=6em, valign=m]{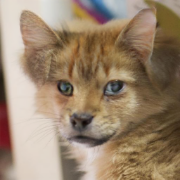}
			\\
			\includegraphics[width=6em, valign=m]{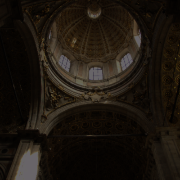}
			& ~
			& ~
			& \includegraphics[width=6em, valign=m]{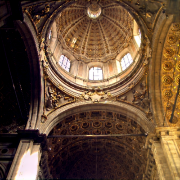}
			& \includegraphics[width=6em, valign=m]{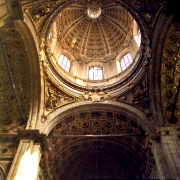}
			& \includegraphics[width=6em, valign=m]{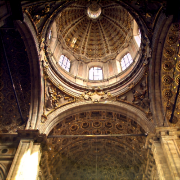}
			& \includegraphics[width=6em, valign=m]{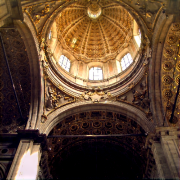}
			\\
			\includegraphics[width=6em, valign=m]{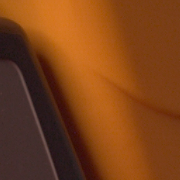}
			& ~
			& ~
			& \includegraphics[width=6em, valign=m]{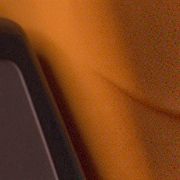}
			& \includegraphics[width=6em, valign=m]{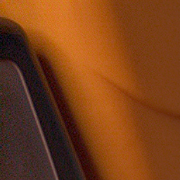}
			& \includegraphics[width=6em, valign=m]{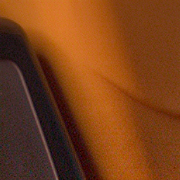}
			& \includegraphics[width=6em, valign=m]{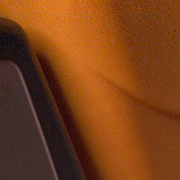}
			\\
			~ \\
			\small \multirow{2}{*}{Input} &
			~ &
			~ &
			\small left low $\mapsto$ &
			\small left high $\mapsto$ &
			\small top low $\mapsto$ &
			\small top high $\mapsto$ 
			\\
			~ &
			~ &
			~ &
			\small right high &
			\small right low &
			\small bottom high &
			\small bottom low
			
		\end{tabular}}
	\caption{More results on pixel-level spatial control. From left to right, we show the input image, continuous pixel-level modulation with intensities changing from left low to right high, left high to right low, top low to bottom high, top high to bottom low respectively}
	\label{fig:MoreSpatial}
\end{figure*}

\end{document}